\documentclass{article}

\usepackage{microtype}
\usepackage{graphicx}
\usepackage{subfigure}
\usepackage{booktabs} 
\usepackage{url}
\usepackage{multicol}
\usepackage{multirow}
\usepackage{wrapfig}
\usepackage{float}
\usepackage{xcolor}
\usepackage{tabularx}
\usepackage{caption}
\usepackage{enumitem}
\usepackage[british]{babel}

\usepackage{amsmath,amsfonts,bm}

\def\eqref#1{equation~\ref{#1}}

\def\1{\bm{1}}

\DeclareMathAlphabet{\mathsfit}{\encodingdefault}{\sfdefault}{m}{sl}
\SetMathAlphabet{\mathsfit}{bold}{\encodingdefault}{\sfdefault}{bx}{n}

\newcommand{\pdata}{p_{\rm{data}}}

\newcommand{\E}{\mathbb{E}}

\newcolumntype{P}[1]{>{\centering\arraybackslash}p{#1}}
\newcommand{\centered}[1]{\begin{tabular}{c} #1 \end{tabular}}
\usepackage{hyperref}

\iftrue
\newcommand{\berivan}[1]{{\color{red}\textbf{BI}: #1}}
\newcommand{\kristy}[1]{{\color{cyan}\textbf{KC}: #1}}
\newcommand{\armin}[1]{{\color{blue}\textbf{AA}: #1}}
\newcommand{\xin}[1]{{\color{purple}\textbf{XZ}: #1}}
\newcommand{\todo}[1]{{\color{green}\textbf{TODO}: #1}}
\fi

\usepackage[accepted]{icml2021}

\begin{document}

\onecolumn
\icmltitle{Neural Network Compression for Noisy Storage Devices}

\begin{icmlauthorlist}
\icmlauthor{Berivan Isik}{to}
\icmlauthor{Kristy Choi}{to}
\icmlauthor{Xin Zheng}{to}
\icmlauthor{Tsachy Weissman}{to}
\icmlauthor{Stefano Ermon}{to}
\icmlauthor{H.-S. Philip Wong}{to}
\icmlauthor{Armin Alaghi}{ed}
\end{icmlauthorlist}

\icmlaffiliation{to}{Stanford University, US}
\icmlaffiliation{ed}{Meta Reality Labs Research, US}

\icmlcorrespondingauthor{Berivan Isik}{berivan.isik@stanford.edu}

\icmlkeywords{Machine Learning, ICML}

\vskip 0.3in



\printAffiliationsAndNotice{}
\begin{abstract}
Compression and efficient storage of neural network (NN) parameters is critical for applications that run on resource-constrained devices. 
Despite the significant progress in NN model compression, there has been considerably less investigation in the actual \textit{physical} storage of NN parameters. Conventionally, model compression and physical storage are decoupled, as digital storage media with error-correcting codes (ECCs) provide robust error-free storage. However, this decoupled approach is inefficient as it ignores the overparameterization present in most NNs and forces the memory device to allocate the same amount of resources to every bit of information regardless of its importance. 
In this work, we investigate analog memory devices as an alternative to digital media -- one that naturally provides a way to add more protection for significant bits unlike its counterpart, but is noisy and may compromise the stored model's performance if used naively. We develop a variety of robust coding strategies for NN weight storage on analog devices, and propose an approach to jointly optimize model compression and memory resource allocation. We then demonstrate the efficacy of our approach on models trained on MNIST, CIFAR-10 and ImageNet datasets for existing compression techniques. Compared to conventional error-free digital storage, our method reduces the memory footprint by up to one order of magnitude, without significantly compromising the stored model's accuracy.
\end{abstract}
\section{Introduction}
\label{intro}
The rapidly growing size of deep neural networks presents new challenges in their storage, computation, and power consumption for deployment in resource-constrained devices \citep{dean2012large,lecun2015deep}. This makes it crucial to compress and efficiently store NN parameters. 
The most commonly used approach is to separate the problem of model compression from physical storage. Reliable digital storage media, fortified by error correcting codes (ECCs), provide nearly error-free storage to users -- this allows researchers to develop model compression techniques independently from the precise characteristics of the devices used to store the compressed weights \citep{model_comp_old_survey, model_comp_new_survey}.
Meanwhile, memory designers strive to create efficient storage by hiding such physical details from users.

Although the decoupled approach enables isolated investigation of model compression and simplifies the problem, it misses the opportunity to exploit the full capabilities of the storage device.
With no context from data, memory systems dedicate the same amount of resources to each bit of stored information. This is suboptimal as NNs tend to exhibit a considerable amount of redundancy in their parameterization \citep{cheng2015exploration,zhou2018non}.
To address this shortcoming, we investigate the joint optimization of NN model compression and physical storage -- specifically, we perform model compression with the additional knowledge of the memory's physical characteristics.
This allows us to dedicate more resources to important bits of data, while relaxing the resources on less valuable bits. 

This joint optimization scheme, however, is cumbersome to implement in practice on digital storage media due to the device's physical characteristics (Sections~\ref{prelim_PCM} and~\ref{prelim_our_setup}).
We instead turn to analog technology -- in particular, phase-change memory (PCM) -- as a more feasible alternative \citep{Joshi2020AccurateDN, PCM_inference}.
Recent studies have demonstrated the promise of end-to-end analog memory systems for storing analog data, such as NN weights, as they have the potential to reach higher storage capacities than digital systems with a significantly lower coding complexity \citep{zarcone2018joint, PMID:32747716, zheng2018error}. 
Yet despite their advantages, analog storage devices are noisy and may corrupt the written input values. 
This presents several key challenges for the compression task. First, the noise characteristic of such memories is a non-linear function of the input value written onto the cell.
Second, slight perturbations of the NN weights from the memory cell may cause the network's performance to plummet \citep{achille2019information}, which is unaccounted for in most NN compression techniques. Thus our objective is to not only minimize the number of memory cells used to store the given NN model (a standard metric named \emph{storage density}), but also preserve the compressed weights' predictive performance.  

Motivated by the above challenges, we draw inspiration from classical information theory and coding theory to develop a framework for encoding and decoding NN parameters to be stored on analog devices \citep{shannon2001mathematical}. 
In particular, our method: (i) leverages existing compression techniques such as pruning \citep{frankle2018lottery, guo2016dynamic} and knowledge distillation (KD) \citep{distillation,polino2018model,xie2020self} to learn a compressed representation of the NN weights; and (ii) utilizes various coding strategies to ensure robustness of the compressed network against storage noise. \footnote{This joint approach is fundamentally different from the problem of preserving the utility of the lossily compressed noisy data \cite{isik2022learning} or vice-versa -- noise-corrupted compressed data.} To the best of our knowledge, this is the first work on NN compression for analog storage devices with \textit{realistic} noise characteristics, unlike previous work that only investigate white Gaussian noise models \citep{fagbohungbe2020benchmarking, zhou2020noisy}. 

In summary, the contributions of our work are as follows:
\begin{enumerate}
    \item We develop a variety of strategies to mitigate the effect of noise and preserve the NN performance on PCM storage devices.
    \item We present methods to combine these strategies with existing model compression techniques.
\end{enumerate}

Empirically, we evaluate the efficacy of our methods on classification tasks with models trained on MNIST, CIFAR-10 and ImageNet datasets and regression tasks with the Neural Radiance Field (NeRF) model \citep{mildenhall2020nerf}; and show that storage density can be increased by $18$ times on average compared to conventional error-free digital storage.
\section{Preliminaries}
\label{prelims}
\subsection{Analog Storage and Phase Change Memory (PCM)}
\label{prelim_PCM}
Most memory technologies utilize continuous physical values (e.g., charge) as a means of data storage. The full continuous storage range is often divided into intervals and used to store discrete values. One extreme that maximizes the device's performance is to allow only two values (high and low) to be written to each memory cell. With this approach, one bit of information can be stored in a memory cell. Storing more values allows more bits to be stored per cell, but also increases the chance of reading an incorrect value from the memory -- this represents a natural trade-off between memory density and probability of error.  
Therefore, storage devices often use error correcting codes (ECCs) in practice to protect data from such memory errors. 

Although digital storage (i.e., storing discrete values) is the dominant paradigm in physical memory technology, it is still possible to store continuous values in memory cells, and read them back as continuous values, albeit with noise. This approach, also known as \emph{analog storage}, has regained attention recently \citep{PMID:32747716} with the emergence of different non-volatile memory (NVM) technologies, such as PCM. NVMs not only retain the stored information even when the power supply is off, but also 
allow efficient storage of multiple bits per cell.
In the extreme case, NVMs can store continuous values \citep{wong2010phase}, making them more efficient than their digital counterparts that require discrete inputs \cite{zarcone2018joint,zheng2018error}. 
\looseness=-1
Among the various NVM technologies, our work focuses on phase-change memory (PCM) technology
because it: (i) has
faster read/write speed and higher endurance than its competitors; 
and  (ii) can enhance chip performance by reducing the cost of data movement thanks to its on-chip integration. 

\begin{figure*}[ht!]
    \centering
        \subfigure[]{\includegraphics[width=.47\textwidth]{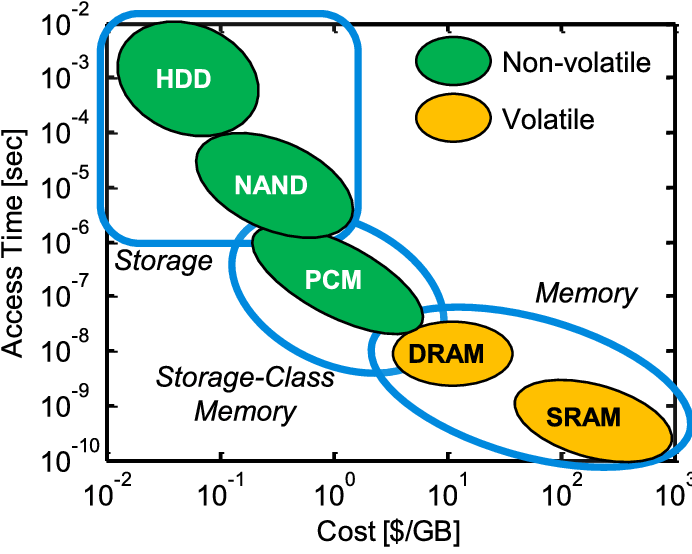}}
        \subfigure[]{ \includegraphics[width=.47\textwidth]{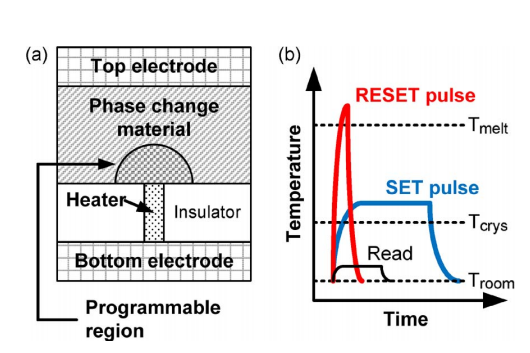}}
    \caption{(a) Figure from \citep{fong2017phase} comparing the access time (read/write time) of various memories versus cost. Significant space has opened between conventional digital memories, namely NAND flash and DRAM, in the memory hierarchy. PCM can fill this gap and further complement the memory hierarchy. (b) The cross-section schematic of PCM cell (left) and RESET and SET pulses are used to program the PCM cell with different temperature (right). Read pulse is used to read the resistance of PCM cell. Figure from \citep{wong2010phase}.}
    \label{fig:PCM_comp}
   \end{figure*}

A major advantage of the PCM memory device over its (digital) competitors is its relatively low cost as compared to its access time (read/write time). 
Commonly used memory devices such as SRAM (static random access memory) and DRAM (dynamic random access memory) have short access time and high endurance, and are thus used as the cache and main memory in a wide range of technologies. 
Storage devices such as NOT-AND (NAND) flash, on the other hand, provide low-cost, high density, and non-volatile storage, but their long access time makes them unsuitable to use as memory devices. 
As shown in Figure~\ref{fig:PCM_comp}(a), PCM successfully bridges the gap between memory and storage devices.
Such advantages of the PCM device allows for utilizing both memory and storage on the main chip, which, when coupled with its high endurance, makes it a compelling alternative to conventional memories such as NAND flash and DRAM.
This is especially true for applications requiring NN inference, where both short access time and non-volatile storage are crucial for real-time deployment.

\subsubsection{PCM Basics: Physics Mechanism, Cell Design and Analog Storage Capability} 
After a significant amount of research in both academia and industry, the PCM array has entered the market as a breakthrough technology bridging the gap between memory and storage \citep{fong2017phase}. 
A common PCM device consists of a phase change material (e.g., GeSbTe) inserted between the top electrode and bottom electrode as in Figure~\ref{fig:PCM_comp}(b)-left.
The information in PCM is stored by utilizing the resistivity difference between the low-conductive amorphous state and high-conductive poly-crystalline states of the phase change material. In the following paragraphs, we briefly explain the physical mechanism by which reading and writing take place on a PCM array.  

To write onto the device, electrical pulses are used to generate phase transformation through joule heating (Figure~\ref{fig:PCM_comp}(b)-right). 
The fresh PCM device is usually in a high-conductive poly-crystalline state. A fast high-temperature (above melting temperature) pulse (RESET pulse) can be used to melt and quench the programming region into low-conductive amorphous states. A longer medium-temperature (above crystallization temperature) pulse, i.e., SET pulse, can then be used to crystallize the programming region back to high-conductive poly-crystalline state. To read from the cell, a smaller pulse is used to measure the resistance of the cell without changing the cell states. 
In particular, the cell state is measured by reading the cell resistance when applying a small bias (read pulse), whose amplitude is small enough not to disturb the cell. 
The cell resistance, interestingly, can be continuously tuned as it is decided by the ratio between amorphous region and polycrystalline region. 
As a result, the PCM device enables analog storage as the cell resistance is an analog value that is determined by the ratio of the two (amorphous and crystalline) phase regions. This property will potentially \emph{increase the storage density of PCM} by storing more than 1 bit per cell.

\subsection{Our Setup}
\label{prelim_our_setup}

\subsubsection{Analog-Storage with 1T1R PCM Array: Measurement Details}

\begin{figure*}[ht!]
    \centering
        \subfigure[1-sigma error bar plot. ]{\includegraphics[width=.32\textwidth]{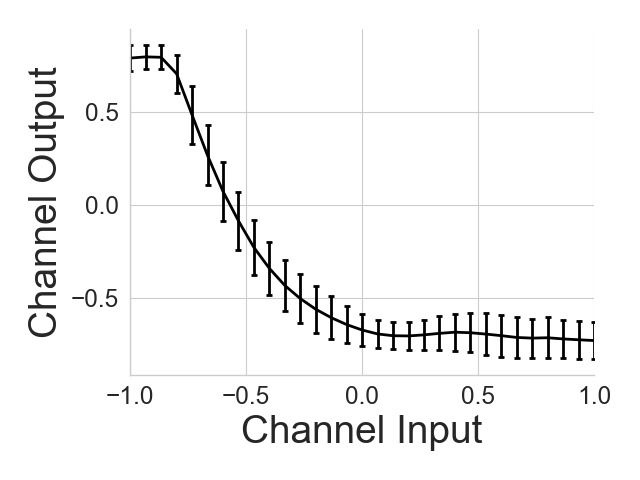}}
        \subfigure[Channel output distribution. ]{\includegraphics[width=.32\textwidth]{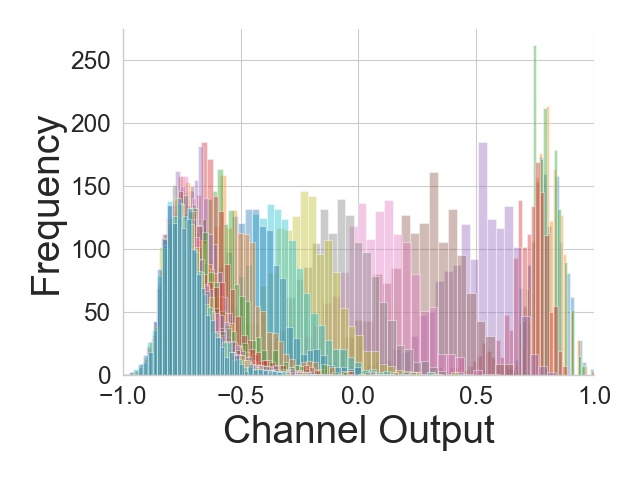}}
        \subfigure[Interpolated measurement data. ]{\includegraphics[width=.32\textwidth]{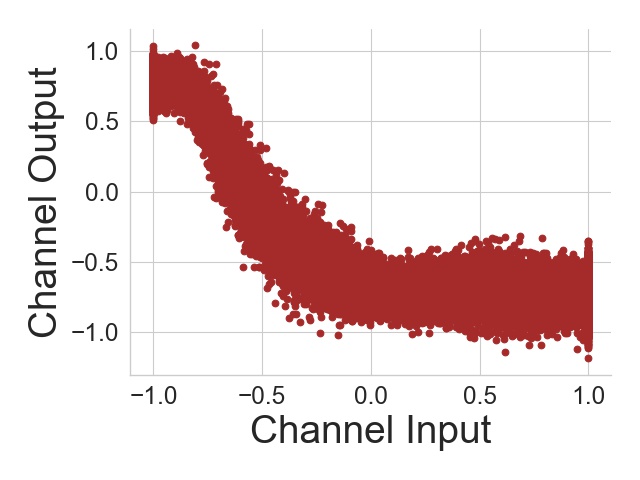}}
    \caption{Characteristics of the channel noise in a PCM cell. (a) 1-standard deviation error bars for each input level, where both the mean and variation of the output has a nonlinear relationship to the input value. (b) Distribution of output values corresponding to 31 distinct input values into the PCM array. (c) Characteristics of the channel noise in a PCM cell, which is dependent on the input value. Each point corresponds to a possible cell output for a given input.}
    \label{fig:channel_app}
   \end{figure*}

To simulate realistic storage, we utilize measurements collected from physical experiments on PCM arrays \cite{wu201840nm} of 1 mega 1T1R cells \citep{wu201840nm}. We use a simple analog programming strategy by first resetting the device to a low-conductive initial state and then setting it with 31 different pulse amplitudes (input levels) and more than 1000 cells for each pulse amplitude. Although this simple programming strategy yields higher noise levels than more complex strategies, the write speed is fast.
The SET pulse amplitude is controlled with a control transistor connected in series to the PCM device (1T1R structure). Figure~\ref{fig:channel_app}(a) shows the 1-standard deviation error bar from cell-to-cell variation for each input level, where both the mean and standard deviation of the output (programmed resistance in log scale) exhibits a non-linear relationship with the input. Figure~\ref{fig:channel_app}(b) is the histogram of output distributions corresponding to the 31 input levels, which shows that the output is roughly Gaussian distributed \emph{conditioned on the input level}. The channel response to the input values in between measurement points is then interpolated in order to construct the \emph{differentiable} continuous analog channel model used in this work as shown in Figure~\ref{fig:channel_app}(c). The differentiability of the model allows for an end-to-end learning scheme (to be discussed in Section~\ref{ae}, see Figure~\ref{fig:appendix_logreg}). Figure~\ref{fig:channel_app}(c) illustrates the PCM model as a \emph{noisy channel}.

Each point in Figure~\ref{fig:channel_app}(c) corresponds to a possible read (output) value for a given write (input) value. For simplicity, we linearly map both the inputs and outputs to be within range $[-1, 1]$.
Our realistic PCM model has different noise mean and standard deviation (std) for each input, which improves over prior works that only investigate white Gaussian noise  \citep{fagbohungbe2020benchmarking, zhou2020noisy} when storing NN models on noisy storage.

\subsubsection{Baselines and Key Assumptions on PCM Storage Modes}
In this section, we establish our two digital storage baselines, as well as the set of assumptions used for our approach. For a fair comparison, we use the same memory device (PCM arrays) in \textit{both digital and analog modes} (Section~\ref{prelim_PCM}) to mitigate confounding factors in the hardware technology.

\paragraph{Ideal baseline for PCM digital storage.} In classical information theory, the maximum amount of data that can be reliably transmitted across a noisy channel (or as in our case, reliably stored into memory cells) with arbitrarily low error is called the \emph{channel capacity} \citep{shannon2001mathematical}. In the case of the PCM noise model shown in Figure~\ref{fig:channel_app}(c), we derive the channel capacity as $2$ bits per cell as in \citep{7047134}, and set this as our \emph{ideal} baseline for digital storage. If we assume 32-bit floating point number representation for each NN weight, then this channel capacity implies that the weight can be reliably stored using 16 digital memory cells. Later in Section~\ref{experiments}, we also consider less precise representations through quantization, which will serve as our stronger baselines.

\paragraph{Practical baseline for PCM digital storage.} Next, we show a more \emph{realistic} baseline for PCM digital storage that is representative of their real-world usage. In practice, various ECCs add extra bits to the data by trading off the overhead coming from additional bits to be stored with the overall reduction in bit error rate (BER). This naturally leads to a rate smaller than the theoretical capacity. We estimate the practically achievable rate of PCM digital storage by multiplying the channel capacity of PCM device ($2$ bits per cell) by a factor $\alpha$, where $\alpha<1$ represents the capacity loss factor (overhead) of practical implementations of Low Density Parity Check (LDPC) codes -- a commonly used ECC in the industry. Specifically, we extract $\alpha = 0.9$ from Figure 6 of \citep{LDPC_design1} that presents LDPC overhead in an additive white Gaussian (AWGN) channel. The practically achievable rate for storing digital data on PCMs would then be $1.8$ bits per cell, smaller than the theoretical limit ($2$ bits per cell). Therefore, a single NN weight (in 32-bit floating point number representation) for this baseline can be reliably stored in 18 digital memory cells. We note that the achieved BER under this assumption is $1e^{-7}$, which is still much larger than the target BER $1e^{-15}$ (digital storage industry standard for acceptable BER). Therefore, 18 cells is actually an optimistic value for the baseline (since the baseline would realistically require more cells in practice), which implies that it is a pessimistic assumption for our method. Nevertheless, we set this as our \emph{realistic} baseline.

\paragraph{PCM analog storage.} In our work, we use PCM cells as \emph{analog} storage devices by storing single-precision floating point data into the device without separating the information into bits. We assume that the read/write circuitry of the PCM has a precision equivalent to that of a single-precision (32-bit) floating point data. This means that a floating point number can be directly written to a memory cell, and the data read back is also a 32-bit float, albeit with the added PCM noise. Although a naive application of PCM analog storage would require more than 1K memory cells to store a single NN weight reliably (while eliminating the effect of the PCM noise), we demonstrate that we can drastically reduce this number to outperform all digital baselines (see Tables~\ref{tab:cifar_sparsity}-\ref{tab:nerf}).

\subsubsection{Additional Assumptions}
In addition to the major assumptions above, we outline a few additional assumptions that are less critical. Our work only focuses on weight storage on the PCM device -- that is, we assume the chip will not be used for training purposes, so there is no need to store gradients onto the cell. We also assume that the pretrained NN activations will be stored in local caches. If such caches are not sufficient for activation storage, then we will have to pay the costly off-chip storage, because PCMs are not efficient for writing purposes (read is cheap; write is expensive).

\subsubsection{Model Limitations}
The PCM model used in this work does not consider second-order effects such as device aging and process variation. The latter can be corrected by adding a compensation term
to our error model, once wafer-level variation data becomes available to us. Furthermore, our method does not consider cell value tuning because its effect was not 
available in our PCM model. Our methodology will not be affected by cell value tuning as its effect will only show up as a new and possibly better error model.

\subsection{Problem Statement and Notation}
To formalize the joint compression problem, we consider a supervised learning setup where $x \in \mathcal{X} \subseteq \mathbb{R}^d$ is the input variable, and $y \in \mathcal{Y} = \{1, \ldots, k\}$ is the label. We assume access to samples $\mathcal{D} = \{(x_i, y_i)\}_{i=1}^n$ drawn from an underlying (unknown) joint distribution $\pdata(x,y)$, which are used to train a NN predictor $f_w: \mathcal{X} \rightarrow \mathcal{Y}$. This network, parameterized by the weights $w \in \mathcal{W}$ to be compressed and stored on analog storage devices (where $\mathcal{W}$ denotes the parameter space of NNs), indexes a conditional distribution  $p_w(y|x)$ over the labels $y$ given the inputs $x$. 

The NN weights $w$ will be exposed to some input dependent device noise $\epsilon(w)$ when written to the analog storage device, yielding noisy versions of the weights $\hat{w}=w+ \epsilon(w)$. 
In our experiments, we find that PCM noise dramatically hurts classification performance -- as a concrete example, the test accuracy of a ResNet-18 model trained on CIFAR-10 drops from 95.10\% (using $w$) to 10\% (using $\hat{w}$) after the weights are corrupted (via naive storage on analog PCM). Thus to preserve NN performance even after it is stored on the analog device, we explore various strategies $g: \mathcal{W} \rightarrow \mathcal{W}$ for designing \textit{reconstructions} of the perturbed weights $g(\hat{w})$ such that the resulting distribution over the output labels $p_{g(\hat{w})}(y|x)$ is close to that of the original network $p_w(y|x)$.

We note that this notion of ``closeness'' between the original weights $w$ and the reconstructed weights $g(\hat{w})$ has several interpretations. In Section~\ref{sparsity_driven}, we explore methods to minimize the distance between $w$ and $g(\hat{w})$ in Euclidean space: 
 \begin{equation*}
 \label{l2}
     \min_{g} \left \Vert g(\hat{w})-w \right \Vert_2.
 \end{equation*}
In Sections~\ref{sensitivity_driven}-\ref{ae}, 
we study ways to minimize the Kullback-Leibler (KL) divergence between these two output distributions:
\begin{equation}
    \label{kl}
     \min_{g} \E _{x \sim p_{\textrm{data}}(x)} [D_{KL}(p_w(y|x) || p_{g(\hat{w})}(y|x))]
\end{equation}
where we \textit{learn} the appropriate transformation $g(\cdot)$.

\section{Robust Neural Network Compression}
We develop several novel methods to provide NNs robustness against storage noise while also minimizing the storage density. In Section~\ref{robust_code}, we describe several robust coding strategies for NN weights to be applied post-training. We exploit the sparsity and sensitivity of NN weights to make our strategies more efficient. In Section~\ref{kl_training}, we propose robust training and robust distillation methods that simultaneously train the NN model to perform well on the downstream task and be robust to storage noise. We achieve this by regularizing the loss function with KL divergence in Eq.~\ref{kl}. Finally, in Section~\ref{ae}, we introduce a probabilistic end-to-end approach to optimize compression and robustness (against storage noise) of the NN model.

\subsection{Robust Coding Strategies}
\label{robust_code}

\begin{figure}[h]
        \centering %
        \subfigure[1 memory cell]{\includegraphics[width=.31\textwidth]{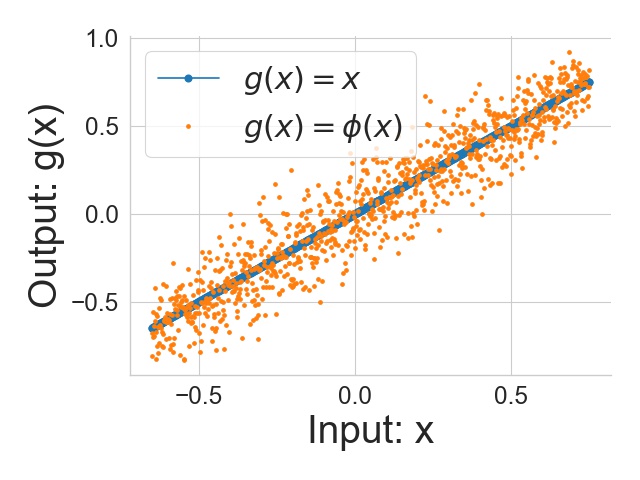}}
        \subfigure[10 memory cells]{\includegraphics[width=.31\textwidth]{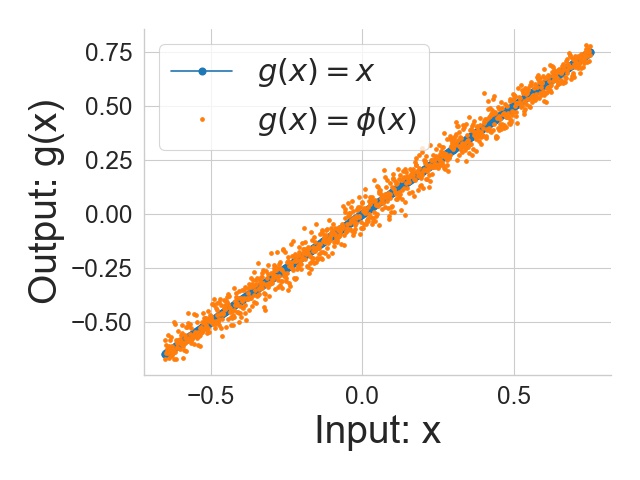}}
        \subfigure[100 memory cells]{\includegraphics[width=.31\textwidth]{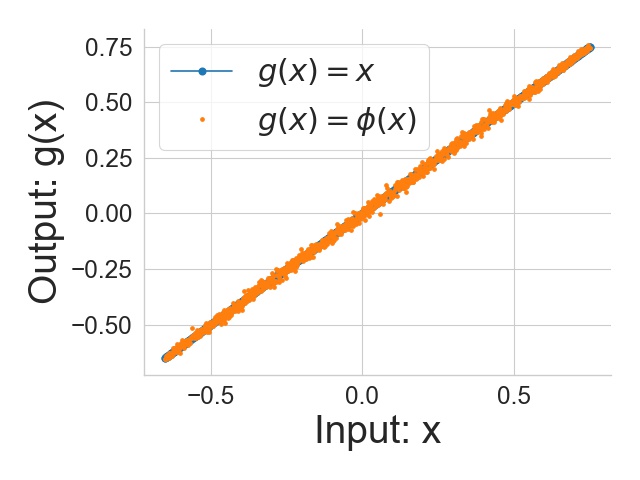}}
    \caption{Behavior of the inverted channel $\phi=C \circ f$ when outputs are read from an average over 1, 10, and 100 memory cells respectively.}\label{fig:channel_inv}
    \vspace{-5mm}
   \end{figure}

In this section, we devise several novel coding strategies for $g(\cdot)$ that can be applied to the model post-training to mitigate the effect of perturbations in the weights. For each strategy, we require a pre-mapping process to remove the input dependence in the mean of the PCM response in Figure~\ref{fig:channel_app}(c).
    \paragraph{Pre-mapping:} Let $C$ represent the PCM channel. The mean function of $C$, denoted as $\mu: \mathcal{X} \rightarrow \mathbb{R}$, in Figure~\ref{fig:channel_app}(c) can be learned via a k-nearest neighbor classifier on the channel response measurements. 
    We can also learn an inverse function $h = \mu^{-1}$ using the measurement data and use it to remove the input dependence in the mean. More precisely, we pre-map the data with $h$ prior to channel usage, which yields an identity function with zero-mean noise $\phi = C \circ h$ (Figure~\ref{fig:channel_inv}), i.e., $\phi(x) = x + \epsilon_0(x)$ where $\epsilon_0(x)$ is a zero-mean noise with input dependent std $\sigma(x)$ due to the PCM channel. 
    Thus the relationship between input weights $w_{in}$ to be stored and output weights $w_{out}$ to be read is:
    \begin{align*}
        w_{out} = \frac{\phi(\alpha \cdot w_{in} - \beta) +\beta}{\alpha},
    \end{align*}
     where $\alpha$ and $\beta$ are scale and shift factors, respectively. Since the noise is zero-mean, we have: 
    \begin{align*}
     w_{out} = \frac{\alpha \cdot w_{in} - \beta +\epsilon_0(w_{in}) +\beta}  {\alpha}= w_{in}+\frac{\epsilon_0(w_{in})}{\alpha}.
     \end{align*}     
 If we use the noisy channel $\phi$ $r$ times (i.e., store the same weight on $r$ ``independent'' cells and average over the outputs -- much like repetition codes in coding theory), the relationship between $w_{in}$ and $w_{out}$ becomes:
\begin{align}
\begin{aligned}
    w_{out}&= \frac{1}{r}\sum_{i=1}^r \left(w_{in}+\frac{\epsilon_{0,i}(w_{in})}{\alpha}\right) \\
    &= w_{in} + \frac{\frac{1}{r}\sum_{i=1}^r \epsilon_{0,i}(w_{in})}{\alpha}
\end{aligned}
\label{noise_eq}
\end{align}
        
Let us define a new random variable $\Tilde{\epsilon}(w_{in}, r) =\frac{1}{r}\sum_{i=1}^r \epsilon_{0,i}(w_{in})$. Notice that the standard deviation of $\Tilde{\epsilon}(w_{in}, r)$  is $\frac{\sigma(w_{in})}{\sqrt{r}}$. Then the standard deviation of $\hat{\epsilon}(w_{in}, r, \alpha)= \frac{\Tilde{\epsilon}((w_{in}, r))}{\alpha}$ in Eq.~\ref{noise_eq} is given by:
$\frac{\sigma(w_{in})}{\alpha \sqrt{r}}.$
More precisely, we have:
\begin{align*}
    w_{out}= w_{in} + \hat{\epsilon}(w_{in}, r,\alpha)
\end{align*}
where the standard deviation of $\hat{\epsilon}(w_{in},r,\alpha)$ is $\frac{\sigma(w_{in})}{\alpha \sqrt{r}}$. This provides us two tools to protect the NN weights against zero-mean noise.  
        \begin{enumerate}[label={}]
            \item \textbf{(Method \#1)} Increase the number of channel usage $r$ (number of PCM cells per weight). 
            \item \textbf{(Method \#2)} Increase the scale factor $\alpha$ under the condition that $\alpha \cdot w_{in}$ satisfies the cell input range limitation ($|\alpha \cdot w_{in}-\beta| \leq 1$). 
        \end{enumerate} 
        
For the first method, we observe in Figure~\ref{fig:channel_inv} that the response becomes less noisy as we increase the number of PCM cells used ($r$).
However, we desire to keep $r$ at a minimum for an efficient storage. 
The second method allows us to make use of the full analog range by scaling the weights. This is particularly useful when storing weights with small magnitude. 
But we cannot increase $\alpha$ without bound because of the device constraint, since 
we must satisfy $-1 \leq\alpha \cdot w_{in} -\beta\leq 1$ (in practice, this constraint is $-0.65 \leq\alpha \cdot w_{in} -\beta \leq 0.75$ since the remaining portion of the response is non-invertible, and therefore unusable for our purposes). Such limitations 
leave more to be desired for a general-purpose robust coding strategy for NN weights. 
  
\subsubsection{Sparsity-Driven Protection}
\label{sparsity_driven}

\begin{figure}[h]
\label{histogram}
        \centering %
        \subfigure[Distribution of weights (ResNet-18 trained on CIFAR-10).]{\includegraphics[width=.35\textwidth]{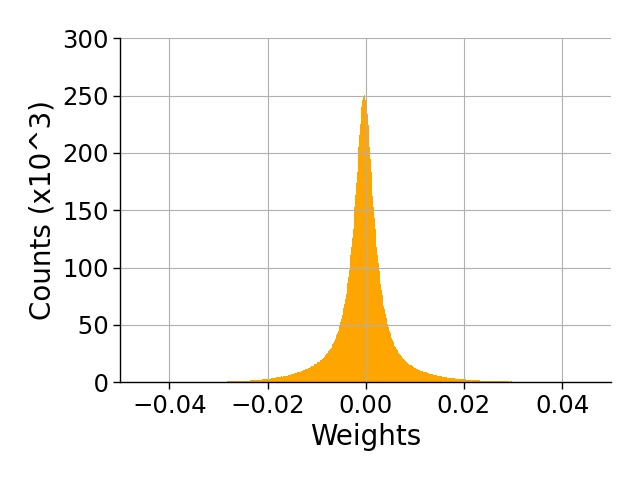}}
        \subfigure[Distribution of sensitivity terms  (ResNet-18 trained on CIFAR-10).]{\includegraphics[width=.35\textwidth]{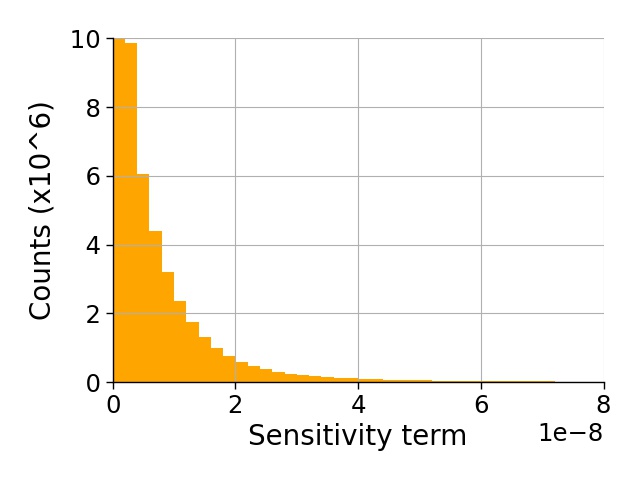}}
        \caption{Histogram of (a) weights of ResNet-18 trained on CIFAR-10, (b) sensitivity terms ($s_j= \frac{1}{N}\sum_{i=1}^N\left(\frac{\partial \log{p_w(y_i|x_i)}}{\partial w_j}\right)^2$ from Section~\ref{sensitivity_driven}) of ResNet-18 trained on CIFAR-10. Since both the weights and sensitivities are sparse, the increase in the average number of cells per weight due to adaptive redundancy and sensitivity-driven protection strategies is negligible. }\label{fig:histogram}
   \end{figure}

Next, we leverage the observation that the weights of a fully-trained NN tend to be sparse \cite{isik2022information} (Figure~\ref{fig:histogram}(a)). 
Table~\ref{tab:cifar_sparsity} shows that 
the test accuracy of a Resnet-18 model on CIFAR-10 with weights corrupted by PCM noise drops from $95.10 \%$ to $10 \%$ (random behavior) when less than $64$ memory cells are used per weight, which does not compare well with our \emph{realistic} baseline for digital storage, where $18$ cells per weight would be enough. Luckily, sparsity-driven strategies help to outperform the \emph{realistic} baseline with an $18 \times$ improvement in the required amount of storage compared to digital storage. 
The 5th row
of Table~\ref{tab:cifar_sparsity} shows that we achieve $95.10 \%$ accuracy with $3$ cells (and $94.44 \%$ even with $1$ cell) per weight using the sparsity-driven protection; on the other hand, the NN performance is compromised without sparsity-driven protection even using $512$ cells per weight with an accuracy of $94.20 \%$ (see \textbf{No Protection} row) -- more than $512 \times$ times reduction in the required amount of storage compared to analog storage without our strategies. 
Figure~\ref{fig:comparison_imagenet} demonstrates the effectiveness of our approach in preserving NN performance on
PCM (Figure~\ref{fig:comparison_cifar_app} for CIFAR-10 in Appendix). We now introduce each sparsity-driven strategies one by one.

    \paragraph{Sign Protection:} When scaling the weights by $\alpha$ to fit them in range $[-0.65, 0.75]$ of $\phi$ (Figure~\ref{fig:channel_inv}), small weights are mapped to values very close to zero. This is problematic because a majority of the trained NN weights have small magnitudes, and thus the NN with reconstructed weights will suffer a severe drop in performance due to sign errors (see Figure~\ref{fig:histogram}(a)). Therefore, we store the sign and magnitude of the weights separately. The sign can be represented by 1 bit. 
    When we store magnitudes instead of actual weights, we can use an $\alpha$ that is twice as large, reducing the variance of noise from Method \#2. 
    The \textbf{No Protection} and \textbf{SP} rows of Table~\ref{tab:cifar_sparsity} illustrate the
    effect of sign protection as the required memory to achieve a test accuracy of $94.1 \pm 0.1 \%$ on CIFAR-10 is reduced by $16 \times$ times (from $512$ cells to $32$ cells).
    
    \paragraph{Adaptive Mapping:} Although protecting the sign bit leads to accuracy gains on CIFAR-10, we still require more than $18$ devices to achieve the original accuracy without PCM noise. 
    To further improve the efficiency, we again use the observation that the majority of nonzero weights are quite small (see Figure~\ref{fig:histogram}(a)). This implies that using different values of $\alpha$ depending on the magnitude of the weight (larger scale factor $\alpha$ for small weights) can reduce the overall variance of the cell's noise from Method \#2. 
    This requires an extra bit to indicate whether a weight is small or large since two different $\alpha$'s are used in encoding and decoding. 
    With this strategy, according to \textbf{SP} and \textbf{SP+AM} rows of Table~\ref{tab:cifar_sparsity}, we can increase the model's accuracy from $9.8\%$ to $90.6\%$ with an average of $1$ cell per weight on CIFAR-10 (and a corresponding increase in accuracy from $0.1 \%$ to $51.3\%$ with an average of $16$ cells per weight for ImageNet).

\paragraph{Adaptive Redundancy:} Finally, we propose to vary the number of PCM cells used for larger and smaller weights. The average number of cells that we aim to minimize is: 
     \begin{align*}
         r_{avg} = \frac{r_{small} \times N_{small} +r_{large} \times N_{large}}{N_{small}+N_{large}}
     \end{align*}
where $N_{small}$ and $N_{large}$ are the number of small and large weights; $r_{small}$ and $r_{large}$ are the number of cells used per weight for small and large weights, respectively. 
Using more cells for larger weights (which are more critical for NN performance) increases the accuracy while it does not increase the average redundancy too much (Method \#1)
since most weights are sparse (see Figure~\ref{fig:histogram}(a)), e.g., $N_{small}=11,168,312$ and $N_{large}=5,650$ in ResNet-18 trained on CIFAR-10. Sign protection and adaptive mapping help protect mostly the small weights while adaptive redundancy protects large weights by reserving more resources for them. As shown in the \textbf{SP+AM} and \textbf{SP+AM+AR} rows of Table~\ref{tab:cifar_sparsity}, combining these strategies
increases the accuracy by $4\% / 66 \%$  on CIFAR-10$/$ImageNet for the 1 cell per weight case without requiring any additional bits.

\subsubsection{Sensitivity-Driven Protection}
\label{sensitivity_driven}
\looseness=-1
While we have demonstrated the success of sparsity-driven protection strategies for $g(\cdot)$ even with $1$ cell per weight for ResNet-18 on CIFAR-10, the method falls short for more complex datasets. For ImageNet, the accuracy of ResNet-50 with sparsity-driven protection against $1$ PCM per weight is $10.6 \%$ lower than the original accuracy without PCM noise ($76.6 \%$, see Table~\ref{tab:cifar_sparsity}). To close this gap, we 
first 
define $\delta_w$ as the final perturbation on the weights, i.e., $g(\hat{w})=w+\delta_w$. 
Then we approximate the KL divergence via a second-order Taylor expansion: 

\begin{align}
    \E _{x \sim p_{\textrm{data}}(x)} [D_{KL}(p_w(y|x) ||  p_{w+\delta w}(y| x))]  \approx \delta w ^T F \delta w + O (||\delta w||^3)
\end{align}
where $F:= \E _{x,y \sim \pdata(x,y)}[\nabla_w \log p_w(y|x) \nabla_w ^T\log p_w(y|x)]$ is the Fisher Information Matrix \citep{grosse2016kronecker, martens2014new}. 
Dropping the off-diagonal entries in $F$ yields 
\begin{align}
    \delta w ^T F \delta w & \approx \E _{x,y \sim \pdata(x,y)} \sum_{j=1}^d \left( \delta w_j \cdot \frac{\partial \log{p_w(y|x)}}{\partial w_j} \right)^2 \\
    & = \sum_{j=1}^d \delta w_j^2 \E _{x,y \sim \pdata(x,y)} \left( \frac{\partial \log{p_w(y|x)}}{\partial w_j} \right)^2
\end{align}
where $d$ is the number of weights. Thus KL divergence between the conditional distribution parameterized by the original ($w$) and perturbed weights ($g(\hat{w})$) that we aim to minimize is: 
\begin{align}
    \E _{x \sim p_{\textrm{data}}(x)} [D_{KL}(p_w(y|x) || p_{w+\delta w} (y|x))] \approx \frac{1}{N}\sum_{j=1}^d \delta w_j^2 \sum_{i=1}^N \left( \frac{\partial \log{p_w(y_i|x_i)}}{\partial w_j} \right)^2
\end{align}

For each weight $w_j$, we can estimate how much performance degradation $\delta w_j$ can cause by evaluating the term 
\begin{align}
s_j = \frac{1}{N}\sum_{i=1}^N \left(\frac{\partial \log{p_w(y_i|x_i)}}{\partial w_j} \right)^2, 
\end{align}
which we call ``sensitivity". 
For the storage of sensitive weights $w_j$ with large $s_j$, we should use more PCM cells as slight perturbations of these weights may lead to significant changes in the output distributions (Method \#1).
This will not significantly affect the number of cells per weight on average ($r_{avg}$) because the gradients of a fully trained network (which correlate with the sensitivity) are known to be sparse. Figure~\ref{fig:histogram}(b) further demonstrates the sparsity of the sensitivity terms. 
We combine sensitivity- and sparsity-driven protection to vary the number of memory cells per weight. 
Table~\ref{tab:cifar_sparsity} demonstrates that
sensitivity-driven protection provides a $2.8 \%$ improvement in test accuracy using $1$ cell per weight for ResNet-50 trained on ImageNet.  Figure~\ref{fig:comparison_imagenet} illustrates that sparsity-driven and sensitivity-driven protection strategies provide $2056\times$ more efficient storage of NN models by preserving the accuracy against PCM noise. 
We refer the reader to Figure~\ref{fig:comparison_cifar_app} in the Appendix for similar results on CIFAR-10. 

\setlength{\tabcolsep}{3pt}
\begin{table}[!h]
\centering
\resizebox{\textwidth}{!}{
\begin{tabular}{clccccccccccc}
\toprule
        &                    & 2056 cells & 512 cells      & 64 cells    & 32 cells    & 16 cells         & 8 cells     & 4 cells   & 3 cells   & 2 cells & 1 cell & Additional Bits\\ \midrule
&No Protection           & 95.1   & 94.2         & 27.1        & 9.0       & 10.2            & 10.2       & 9.9      & 9.6       & 9.8   & 10.3  & 0\\
&SP                      &95.1 & 95.0 & 94.2                            & 94.0       & 92.8            & 89.5       & 67.0     & 41.9     & 11.8   & 9.8 & 1\\ 
ResNet-18&AM+AR                   & 95.0 & 95.0 & 94.8                             & 94.7      & 94.4             & 93.7       & 93.1   & 92.7      & 89.2   & 58.0 & 1\\
CIFAR-10&SP+AM                   &95.1 & 95.1 & 95.0                            & 95.0       & 94.8            & 94.7       & 94.6     & 93.9     & 93.2   & 90.6  & 2\\ 
&SP+AM+AR                &95.1 & 95.1 & 95.0                        &\textbf{95.0} & \textbf{95.0} & \textbf{95.0}   & \textbf{95.0}  & \textbf{95.1}    & \textbf{94.8}  & \textbf{94.4}    & 2\\
&SP+AM+AR+Sens.    & 95.1 & 95.1 & 95.1                            & \textbf{95.1}    & \textbf{95.1}  & \textbf{95.1}  & \textbf{95.0}   & \textbf{95.1}  & \textbf{94.8}  & \textbf{95.0} & 3\\ \midrule
&No Protection               &  67.6      & 0.1 & 0.1  & 0.1       & 0.1       & 0.1           & 0.1       & 0.1  & 0.1  & 0.1  & 0\\
&SP                          & 75.8       & 4.2          & 1.1       & 0.1       & 0.1           & 0.1       & 0.1      & 0.1    & 0.1     & 0.1  & 1\\ 
ResNet-50&AM+AR                          &  76.1     &   76.0        &  70.2      &    70.0    &    67.8        &  65.5      & 46.1      & 35.8    & 10.3  & 0.1  & 1\\ 
ImageNet&SP+AM                       & 76.6      & 75.5           & 75.0        & 68.2        & 65.0            & 50.2        & 48.8   &  25.1       & 12.2     & 1.0  & 2\\
&SP+AM+AR                    & \textbf{76.6} & \textbf{76.6} & \textbf{76.6}  & \textbf{76.6}       & \textbf{76.6}   & \textbf{76.4}  & \textbf{75.9}    & \textbf{76.0}  & \textbf{75.8} & \textbf{75.5} & 2\\
&SP+AM+AR+Sens.     & \textbf{76.6}  & \textbf{76.6} & \textbf{76.6}     & \textbf{76.6}       & \textbf{76.5}   & \textbf{76.6}  & \textbf{76.2}    & \textbf{75.9}  & \textbf{76.0} & \textbf{75.9} & 3 \\\midrule
&No Protection               &  34.5      & 0.1 & 0.1  & 0.1       & 0.1       & 0.1           & 0.1       & 0.1  & 0.1  & 0.1  & 0\\
&SP                          & 69.9       & 2.0          & 0.1       & 0.1       & 0.1           & 0.1       & 0.1      & 0.1    & 0.1     & 0.1  & 1\\ 
MobileNet-v2&AM+AR          &  71.4     &   68.4        &  65.1      &    61.0    &    52.3        &  36.8      & 22.6      & 11.9    & 0.1  & 0.1  & 1\\ 
ImageNet&SP+AM                       & 71.8      & 71.2           & 70.6        & 65.4        & 50.5            & 43.8        & 32.0   &  16.7       & 0.1     & 0.1  & 2\\
&SP+AM+AR                    & \textbf{71.8} & \textbf{71.8} & \textbf{71.8}  & \textbf{71.8}       & \textbf{71.8}   & \textbf{71.4}  & \textbf{71.0}    & \textbf{70.5}  & \textbf{70.2} & \textbf{69.5} & 2\\
&SP+AM+AR+Sens.     & \textbf{71.8}  & \textbf{71.8} & \textbf{71.8}     & \textbf{71.8}       & \textbf{71.8}   & \textbf{71.6}  & \textbf{71.2}    & \textbf{71.1}  & \textbf{71.0} & \textbf{70.4} & 3 \\\midrule
&No Protection               & 39.8      & 0.1 & 0.1  & 0.1       & 0.1       & 0.1           & 0.1       & 0.1  & 0.1  & 0.1  & 0\\
&SP                          & 72.4       & 5.2          & 0.4       & 0.1       & 0.1           & 0.1       & 0.1      & 0.1    & 0.1     & 0.1  & 1\\ 
EfficientNet-B2&AM+AR          &  79.6     &   77.1        &  72.8      &    64.7    &    49.3        &  37.2      & 17.5      & 9.2    & 0.1  & 0.1  & 1\\ 
ImageNet&SP+AM                       & 80.2      & 78.6           & 76.5        & 69.8        & 57.0            & 40.8        & 24.1   &  15.9       & 0.1     & 0.1  & 2\\
&SP+AM+AR                    & \textbf{80.2} & \textbf{80.2} & \textbf{71.8}  & \textbf{80.2}       & \textbf{79.9}   & \textbf{79.2}  & \textbf{78.6}    & \textbf{78.0}  & \textbf{77.4} & \textbf{76.4} & 2\\
&SP+AM+AR+Sens.     & \textbf{80.2}  & \textbf{80.2} & \textbf{80.2}     & \textbf{80.1}       & \textbf{80.2}   & \textbf{80.1}  & \textbf{79.8}    & \textbf{79.2}  & \textbf{78.8} & \textbf{77.9} & 3 \\
\bottomrule
\\
\end{tabular}
}
\caption{Accuracy of ResNet-18 on CIFAR-10; and ResNet-50, MobileNet-v2, Efficientnet-B2 on ImageNet when weights are perturbed by the PCM cells. Baseline accuracy (without noise) is $95.1 \%$ for ResNet-18 on CIFAR-10, $76.6 \%$ for ResNet-50 on ImageNet, $71.8 \%$ for MobileNet-v2 on ImageNet, and $80.2 \%$ for EfficientNet-B2 on ImageNet. SP: sign protection, AM: adaptive mapping, AR: sparsity-driven adaptive redundancy, Sens.: sensitivity-driven adaptive redundancy. Results are averaged over three runs. Higher is better. We provide the average number of cells, including the ones required to store the additional bits, in Table~\ref{tab:num_cell_accuracy}. 
}
\label{tab:cifar_sparsity}
\end{table}

\setlength{\tabcolsep}{7pt}
\begin{table}[h]
\resizebox{\textwidth}{!}{
\begin{tabular}{lcccccc}
\toprule
&                           SP+AM+AR                   & SP+AM+AR                       & SP+AM+AR                       & SP+AM+AR+Sens.      & SP+AM+AR+Sens. & SP+AM+AR+Sens.  \\ 
&                           4 cells                    & 3 cells                        & 2 cells                        & 4.5 cells           & 3.5 cells      & 2.5 cells       \\ \midrule
ResNet-18 on CIFAR-10       &  95.1                    & 94.8                           & 94.4                           & 95.1                & 94.8           & 95.0  \\
ResNet-50 on ImageNet       &  76.0                    & 75.8                           & 75.5                           & 75.9                & 76.0           & 75.9  \\
MobileNet-v2 on ImageNet    &  70.5                    & 70.2                           & 69.5                           & 71.1                & 71.0           & 70.4 \\
EfficientNet-B2 on ImageNet &  78.0                    & 77.4                           & 76.4                           & 79.2                & 78.8           & 77.9 \\
\bottomrule
\\
\end{tabular}}
\caption{A sample of accuracy vs. average number of cells required to store: 1) the continuous weight values and 2) the additional bits. Numbers taken from Table~\ref{tab:cifar_sparsity} by computing the average number of cells considering the ones that are required to store the additional bits. Baseline accuracy (without noise) is $95.1 \%$ for ResNet-18 on CIFAR-10, $76.6 \%$ for ResNet-50 on ImageNet, $71.8 \%$ for MobileNet-v2 on ImageNet, and $80.2 \%$ for EfficientNet-B2 on ImageNet. SP: sign protection, AM: adaptive mapping, AR: sparsity-driven adaptive redundancy, Sens.: sensitivity-driven adaptive redundancy. Results are averaged over three runs. Higher is better. 
}
\label{tab:num_cell_accuracy}
\end{table}

\subsection{KL Regularization for Robustness}
\label{kl_training}
The following set of techniques for constructing $g(\cdot)$ are designed to correct the errors that the robust coding strategies in the previous section fail to address. 

\label{robust_train}
\paragraph{Robust Training:} For robust training, we regularize the standard cross-entropy loss with 
the KL divergence in Eq.~\ref{kl}.
Specifically, the loss function is as follows:
\begin{align}
     \mathcal{L}(w) &=\E _{x,y \sim \pdata} [-\log p_w(y|x)] + \lambda \E _x [D_{KL}(p_w(y|x) || p_{g(\hat{w})}(y|x))].
     \label{eq:robust_training}
 \end{align}
In our experiments, we add a noise (with a carefully adjusted standard deviation) as a perturbation $\delta w$ (i.e., $g(\hat{w})=w+\delta w$ is a noisy weight) during robust training, 
and we observe that the trained network is more robust to PCM noise and also to pruning effects. This is because the final perturbation $\delta w$ on the weights ($g(\hat{w})=w+\delta w$) can be thought of as a noise or the effect of pruning on the $w$. Although our framework does not involve a pruning step (pruning can be performed independently as we show in Section~\ref{sec:exp_sparsity}), we believe that making NN weights more robust to pruning effects is an important additional feature of our strategy. In particular, when we apply pruning, we have:
$\delta w_j = 0$ for a non-pruned weight $w_j$, and $\delta w_j = -w_j$ for a pruned weight $w_j$. Recall that in magnitude pruning, only the small weights are set to zero, that is $\delta w_j$ is equal to $0$ (for large weights) or $-w_j$ (for small weights). In other words, $\delta w_j$ is always a small value, just like a noise, therefore this strategy could provide robustness against pruning effects as well. 
Our experiments on CIFAR-10 and ImageNet verify that robust training has a protective effect against PCM noise where robust coding strategies are not enough.     

\paragraph{Robust Distillation:} 
Distillation is a well-explored NN compression technique \citep{distillation}. The idea is to first train a teacher network to capture a smooth probability distribution on labels, and then train a smaller student network by distilling the output probability information from the teacher. 
We use distillation to optimize a compressed model to be robust to PCM noise via the noisy student loss:
\begin{align}
     \mathcal{L}_s(w_s) & =(1-\lambda)\E _{x,y \sim \pdata} [-\log p_{\hat{w}_s}(y|x)]  + \lambda \E _x [D_{KL}(p_{w_t}(y|x) || p_{g(\hat{w}_s)}(y|x))]
     \label{eq:robust_distill}
 \end{align}
where $\lambda \in [0,1]$ is a scalar, $w_t$ and $w_s$ are teacher and student weights, $g(\hat{w}_s)= g(w_s+\epsilon(w_s))=w_s+\delta w_s$ with $\epsilon(w_s)$ being the PCM noise and $\delta w_s$ being the noise injected onto the student network's weights. 
Although \citep{zhou2020noisy} provides an initial exploration into distillation for noisy storage, 
we leverage experimental data collected from real storage devices to build a realistic model of the PCM noise.

\subsection{End-to-End Learning}
\label{ae}
Finally, we explore a probabilistic, end-to-end learning approach for $g(\cdot)$ that jointly optimizes over the model compression and the known characteristics of the noisy storage device. We assume access to a set of weights $\{W_j\}_{j=1}^K$ from $K$ models that have been trained on the same dataset $\mathcal{D}$ -- this serves as an empirical distribution over NN weights, as different initializations of NNs typically converge to different local minima. Additionally, we assume that each weight is a sample from a normal distribution $W_{ij} \sim \mathcal{N}(0,1)$ \citep{dziugaitetowards, zhou2018non}. \\

To learn the representation of the NN weight, we maximize the mutual information (MI) between the original network weight $W$ and the compressed weight $Z$ that has been corrupted by the noisy channel. Following \cite{choi2019neural}, our coding scheme can be represented by the following Markov chain\footnote{Here, we introduce the new notation $\hat{Z}$ and $Z$ for the compressed and noisy version of the weights because, in the end-to-end learning approach, we
have neural encoder and decoder networks to compress ($W \rightarrow \hat{Z}$) and decompress ($Z \rightarrow \hat{W}$) the weights. This notation was not necessary in the previous sections, as we did not have such an encoder-decoder pair.}: 
\begin{align}
W \rightarrow \hat{Z} \rightarrow Z \rightarrow \hat{W},
\end{align}
where $\hat{Z}$ denotes the compressed weight and $Z = \hat{Z} + \epsilon(Z)$ where $\epsilon(Z)$ denotes the input-dependent PCM noise. Then, we obtain the following lower bound to the true MI:
\begin{align}
\label{mi_bound}
    I(W;Z) &= H(W) - H(W|Z) = -H(W|Z) + \text{const.} \\
    & \geq \mathbb{E}_{q_\phi(Z|W)}[\log p_\theta(W|Z)]
\end{align}
where $q_\phi(Z|W)$ and $p_\theta(W|Z)$ denote approximations to the true posteriors $p(Z|W)$ and $p(W|Z)$, respectively \citep{barber2003algorithm}. We train an autoencoder $g_{\phi,\theta}(\cdot)$ with a stochastic encoder and decoder 
to learn these variational posteriors. The decoder is trained to output a set of reconstructed weights such that its predictions are close to those of the original network. Notably, our approach differs from KD in that we \textit{learn} the weight compression scheme rather than using a fixed student network architecture.
\section{Experimental Results}
\label{experiments}
We empirically investigate: (1) whether our protection strategies for $g(\cdot)$ help to preserve NN accuracy; and (2) the improvements in storage efficiency when using our approach. All experimental results are averaged over 3-5 runs. For conciseness, we report the average and refer the reader to Appendix~\ref{confidence} for the complete results.

For all the classification experiments, we consider models trained on three image datasets: MNIST \citep{lecun2010mnist}, CIFAR-10 \citep{krizhevsky2009learning}, and ImageNet \citep{imagenet_cvpr09}. We use the standard train/val/test splits for MNIST and CIFAR-10 datasets and the standard train/val split for the ImageNet dataset. 
For MNIST, we use two architectures: LeNet \citep{lecun1998gradient}, and a 3-layer MLP. For CIFAR-10, we use two types of ResNets \citep{he2016deep}: ResNet-18 and a slim version of ResNet-20. For ImageNet, we use pretrained ResNet-50 from PyTorch \citep{he2016deep}, and lightweight models MobileNet-v2 \citep{sandler2018mobilenetv2} and EfficientNet-B2 \citep{tan2019efficientnet}. For the regression experiment, we use the standard Neural Radience Fields (NeRF) model \citep{mildenhall2020nerf} on the fern dataset (\url{https://github.com/bmild/nerf}).  For additional details on architectures and hyperparameters, we refer the reader to Appendix~\ref{hyperparams}.




\subsection{Sparsity and Sensitivity Driven Protection} \label{sec:exp_sparsity} We show the effect of sparsity- and sensitivity-driven protection in Figure~\ref{fig:comparison_imagenet} on ResNet-50 for ImageNet (Figure~\ref{fig:comparison_cifar_app} for CIFAR-10 in Appendix). The exact numerical results are given in Table~\ref{tab:cifar_sparsity}. 

In CIFAR-10 experiments, the number of small weights was $N_{small}=11$M and the number of large weights was $N_{large}=5.6$K and number of cells per weight on average for adaptive redundancy is not more than $0.02$ above the listed numbers in the table. Similarly, in ImageNet experiments, $N_{small}=25.4$M, $N_{large}=15$K and the difference between the number of cells per weight on average and the listed number is always smaller than $0.06$.

In Table~\ref{tab:cifar_sparsity}, we provide detailed experimental results on the effect of each sparsity- and sensitivity-driven strategy with ResNet-10 on CIFAR-10; and ResNet-50, MobileNet-v2, and EfficientNet-B2 on ImageNet. We provide the required number of cells per weight to store the continues value of the weight (3rd-12th columns) and the additional bits to store for each strategy (last column) separately. In Table~\ref{tab:num_cell_accuracy}, we report some of the results from Table~\ref{tab:cifar_sparsity} again, this time by providing the number of cells per parameter required to store 1) the continuous weight value and 2) the additional bits. Recall that we can store $2$ bits or $1.8$ bits digitally in one PCM cell with the \emph{ideal} or \emph{realistic} baselines, respectively. It is seen from the two tables that sparsity driven protection strategies are enough for ResNet-18 on CIFAR-10 to preserve the classification accuracy ($95.10 \%$)  with $4$ cells per weight. Moreover the accuracy  with $2.5$ cells per weight. is only $0.1 \%$ less than the baseline accuracy ($95.10 \%$). Similar observation can be made for the ImageNet results as well. 
Table~\ref{tab:cifar_sparsity} also shows that sign protection is a critical step: for instance, the \textbf{AM+AR} and \textbf{SP+AM+AR} rows indicate that sign protection can increase the accuracy from $58 \%$ \textbf{AM+AR} to $94.4 \%$ \textbf{SP+AM+AR} on CIFAR-10.\footnote{We would like to note that it is possible to improve the performance of the noisy NNs further by  retraining them after the weights are read from the PCM cells. In fact, our experiments suggest that for the models stored with SP+AM+AR and SP+Am+AR+Sens. strategies using $1$ cell per weight, retraining recovers the original accuracy in $1-3$ epochs. However, we believe it is not realistic to assume that the stored NNs can be retrained further since we are particularly interested in resource-constrained edge devices.}

\begin{figure}[h]
\centering
\includegraphics[width=.47\textwidth]{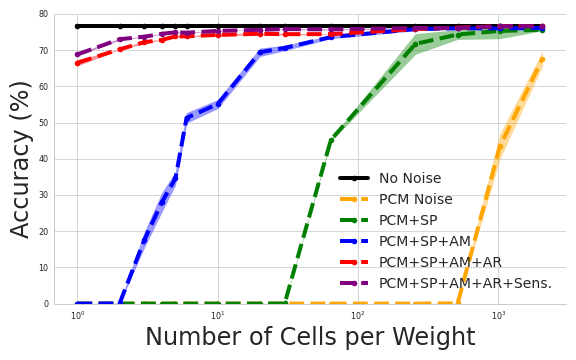}
\caption{ResNet-50 on ImageNet. SP: sign protection, AM: adaptive mapping, AR: sparsity-driven adaptive redundancy, AP: adaptive protection (AM+AR), Sens.: sensitivity-driven adaptive redundancy.  } \label{fig:comparison_imagenet}
\end{figure}

We also combine sparsity driven protection with pruning.
Results on 
$90 \%$ pruned ResNet-18 (on CIFAR-10) are given in Table~\ref{tab:cifar_pruning_app} in Appendix~\ref{pruning_app}. We follow the standard pruning procedure: first train the model as usual, then prune $90\%$ of the weights, and then retrain the remaining non-pruned weights by keeping the pruned weights frozen at value zero. In our experiment, we retrain the pruned model for $20$ epochs. Then for the storage of this pruned model, we assume that the pruning mask could be stored using the compressed sparse row (CSR) or compressed sparse column (CSC) format and Huffman coding, as detailed in \citep{deep_compression}. This would require at most $0.5$ bits per weight. Then we need to store the continuous weight values of the \emph{non-pruned weights}. Note that this is the standard technique for storing a pruned network: (1) first store the pruning mask, (2) then store the values of the non-pruned weights. In Table~\ref{tab:cifar_pruning_app}, the number of cells we report are computed as follows:
\begin{align*}
 \text{avg. \# of cells per weight} = \frac{\text{\# of pruned weights}}{ 
\text{total \# of weights}} \times (\text{avg. \# of cells per non-pruned weight}) + 0.25,
\end{align*}
translating the number of cells per non-pruned weight to number of cells per (pruned or non-pruned) weight and also considering additional $0.25$ cells per parameter to store the pruning mask.
Note that the accuracy of the pruned model without noise is $94.8 \%$ and we get $94.2 \%$ test accuracy with PCM noise using only $1.35$ cells per weight. 
Compared to the \emph{realistic} digital baseline (no compression and each $32$-bit weight is digitally stored on $18$ PCM cells), we provide $\frac{18}{1.35} = 13.3\times$ more efficient storage with analog storage combined with our strategies and $90\%$ pruning.

\subsection{Robust Training and Distillation}

We compare robust training with naive training (training with no noise) in Table~\ref{tab:cifar_robust_train} on ResNet-18 (on CIFAR-10) and on ResNet-50 (on ImageNet). In our experiments, we use $\lambda = 0.5$ as the coefficient of the KL term in the loss. We experimented with different combinations of the sparsity and sensitivity driven strategies, and in all cases, robust training provides better robustness against PCM noise than the naive training. Robust training provides robustness against pruning as well. When ResNet-18 trained without noise is pruned with $90\%$ sparsity, the accuracy drops from $95.1 \%$ to $\mathbf{90.2 \%}$. However, the same model trained with $N(0,0.006)$ gives $\mathbf{95.0 \%}$ accuracy after $90\%$ pruning.

\setlength{\tabcolsep}{7pt}
\begin{table}[h]
\begin{tabular}{llcccc}
\toprule
&                            & Naive Train       & Robust Train                       & Robust Train      & Average \\ 
&                            &(no noise)             &  (with $\sigma = 0.01$)                    & (with $\sigma =0.006$)   &  Number of Cells \\ \midrule
&No Noise                    & 95.10                 & 95.50                                 & \textbf{95.60}                    & - \\
&PCM (No Protection)   & 9.70     & 8.30                   & 9.90                    &1\\
&PCM+SP                & 9.70      & 10.63                    & 10.33                    & 1.5 \\
CIFAR-10&PCM+AP                & 27.69    & \textbf{86.20}            & 81.83                    & 1.5\\
&PCM+SP+AP             & 90.60     & 94.73                 & \textbf{94.80}                    & 2 \\
&PCM+AP+Sens           & 36.21    & \textbf{86.73}          & 78.93                     & 2\\
&PCM+SP+AP+Sens        & 94.95     & \textbf{95.03}            & \textbf{95.03}                   & 2.5\\ \midrule
    
&No Noise                    & 76.60                 & 76.60                                 & 76.60        & - \\
&PCM (No Protection)   & 0.1                 & 0.1                                  & 0.1                  & 1\\
&PCM+SP                & 0.1                 & \textbf{0.4}                        & 0.2                 & 1.5\\
ImageNet&PCM+AP                & 0.1                 & \textbf{0.002}                        & 0.1                 & 1.5\\
&PCM+SP+AP             & 75.50                 & 77.10                                 &  \textbf{77.80}       &2\\
&PCM+AP+Sens           & 0.3                 & \textbf{0.6}                            & 0.5                 &2\\
&PCM+SP+AP+Sens        & 75.90                 & 76.20                    & \textbf{76.50}        & 2.5\\
\bottomrule
\\
\end{tabular}
\caption{Robust training vs. naive training for ResNet-18 on CIFAR-10 and ResNet-50 on ImageNet
. SP: sign protection, AP: adaptive protection (adaptive mapping+sparsity-driven adaptive redundancy), Sens.: sensitivity-driven adaptive redundancy. 
The average number of cells per weight to store the continuous weight value is $1$ in all experiments. The reported average number of cells in the rightmost column includes 1) the number of cells required to store the continuous weight value and 2) the number of cells required to store the additional bits. 
}
\label{tab:cifar_robust_train}
\end{table}

Table~\ref{tab:cifar_distill} shows the distillation results on ResNet-20 distilled from ResNet-18 (on CIFAR-10) with PCM noise applied at test time. We compare three networks: (1) a student ResNet-20 distilled without noise injection, (2) a student ResNet-20 distilled with Gaussian noise injection, and (3) a teacher ResNet-20 (a baseline) trained without noise. As shown in Table~\ref{tab:cifar_distill}, student ResNet-20 distilled with noise injection outperforms both teacher ResNet-20 and student ResNet-20 distilled without noise when weights are perturbed by PCM at test time. (see Appendix~\ref{distill_exp_app} for additional results.) 
\setlength{\tabcolsep}{3pt}
\begin{table}[h!]
\centering
\begin{tabular}{ccccccc}
\toprule
                            & Avg. \# of Cells & Teacher                       & Teacher                               & Student                       & Noisy Student        & Avg. \# of Cells    \\ 
                            & for Cont. Weights   & ResNet-18                     &  ResNet-20                    & ResNet-20            & ResNet-20    & in Total                     \\ \midrule
No Noise                    & - & 95.70                         & 92.50                                  & 92.90                        & \textbf{93.00}     &- \\ \midrule
PCM+AP               &3   & 16.23     & 48.38             & 73.38                   & \textbf{81.75}        & 3.5\\ \midrule
\centered{PCM+SP+AP} & \centered{1 \\ 3} & \centered{93.35 \\ 94.78} & \centered{86.30 \\ 89.73} & \centered{88.58 \\ 89.98} & \centered{\textbf{90.65} \\ \textbf{91.33}} & \centered{2 \\ 4} \\ \midrule
\centered{PCM+AP+Sens.}           & \centered{1} & \centered{9.60}     & \centered{29.68}         & \centered{38.18}                  & \centered{\textbf{69.49}}       &\centered{2} \\ \midrule

\centered{PCM+SP+AP+Sens.} & \centered{1 \\ 3} & \centered{93.36 \\ 94.90} & \centered{88.40 \\ 89.92} & \centered{88.96 \\ 90.44} & \centered{\textbf{91.10} \\ \textbf{91.78}} & \centered{2.5 \\ 4.5} \\
\bottomrule
\\
\end{tabular} 
\caption{Accuracy of ResNet-20 distilled from ResNet-18 on CIFAR-10. The average number of cells per weight to store the continuous weight value is given in the leftmost column. The reported average number of cells in the rightmost column includes 1) the number of cells required to store the continuous weight value and 2) the number of cells required to store the additional bits. }
\label{tab:cifar_distill}
\end{table}

\subsection{Analog-Digital Comparison} 
We also consider quantization as a way to improve the efficiency of digital storage. Table~\ref{tab:analog_digital} shows a comparison between analog storage improved with our strategies and digital storage improved with quantization techniques. We consider both the \emph{ideal} ($2$ bits per cell) and 
\emph{realistic} baselines ($1.8$ bits per cell). For digital storage, we apply quantization using different techniques from the literature \citep{ jacob2018quantization, banner2019post}.  
We find the number of cells to store one quantized weight in both \emph{ideal} and \emph{realistic} cases. Then we adjust the number of cells used to store one weight in analog PCMs to be the same as the digital PCMs for both baselines, by adapting the redundancy for large and less sensitive weights. For instance, when \citep{banner2019post} performs 4-bit quantization, each parameter is represented by $4$ bits. In the \emph{ideal} baseline, this would require $2$ cells per parameter. We adjust the parameters in our robust strategies so that the number of cells required to store one parameter is $2$ and report the result ($76.02$) in the "PCM (analog) + our robust strategies" row under "\emph{Ideal}" column. We repeat the same procedure for the \emph{realistic} baseline which requires $2.22$ cells per parameter to store $4$ bits. We report the result ($76.08$) in the "PCM (analog) + our robust strategies" row under "\emph{Realistic}" column.
As shown in Table~\ref{tab:analog_digital}, 
noisy analog storage improved by our strategies outperforms digital storage of quantized weights. We do not compare against more aggressive quantization techniques \citep{scalableQuant,  choi2020universal, QuantNoise, adaptiveQuant, oktay2019scalable, DeepCABAC} that can achieve higher efficiency in digital storage since they incur a huge complexity with multiple retraining stages.

\setlength{\tabcolsep}{7pt}
\begin{table}
\begin{tabular}{llcc}
\toprule
&   &  \emph{Ideal}                       & \emph{Realistic}\\ \midrule
&PCM (digital) + 8-bit quantization with \citep{jacob2018quantization}   & 68.30 & 68.30 \\
$\sim 4$ cells per parameter & PCM (digital) + 8-bit quantization with ACIQ \citep{banner2019post} & 73.60 & 73.60 \\
&PCM (analog) + our robust strategies  & \textbf{76.02} & \textbf{76.08} \\ \midrule
    
&PCM (digital) + 4-bit quantization with \citep{jacob2018quantization}   & 72.50 &  72.50 \\
$\sim 2$ cells per parameter &PCM (digital) + 4-bit quantization with ACIQ \citep{banner2019post}  & 73.80 & 73.80\\
&PCM (analog) + our robust strategies  & \textbf{75.50} & \textbf{75.62} \\
\bottomrule
\\
\end{tabular}
\caption{Digital vs. analog storage for ResNet-50 on ImageNet. For digital storage, number of cells is reduced via 4-bit and 8-bit quantization techniques \citep{ jacob2018quantization, banner2019post}. For analog storage, it is reduced via our strategies. For fair comparison, the number of PCM cells per weight in analog storage is adjusted to be the same as the digital storage (in both \emph{ideal} and \emph{realistic} baselines). For instance, when \citep{banner2019post} performs 4-bit quantization, each parameter is represented by $4$ bits. In the \emph{ideal} baseline, this would require $2$ cells per parameter. We adjust the parameters in our robust strategies so that the number of cells required to store one parameter is $2$ and report the result ($76.02$) in the "PCM (analog) + our robust strategies" row under "\emph{Ideal}" column. We repeat the same procedure for the \emph{realistic} baseline which requires $2.22$ cells per parameter to store $4$ bits. We report the result ($76.08$) in the "PCM (analog) + our robust strategies" row under "\emph{Realistic}" column.}
\label{tab:analog_digital}
\vspace{-2mm}
\end{table}

\subsection{Regression Models}
We test our strategies on a regression setting as well. For the regression task, we consider the Neural Radiance Fields (NeRF) framework \citep{mildenhall2020nerf} that contains a neural network to model the relation between 1) the 3D coordinates of a point in a given 3D scene and the view direction; and 2) the view-dependent continuous color\footnote{Color is then discretized during evaluation.} and volume density at that 3D point. More precisely, a NeRF model takes 3D positions $(x,y,z)$ of a point in 3D space and a particular direction to view the 3D scene $(\theta, \phi)$; and outputs a view-dependent RGB color and a volume density at that point. We note that this is a nontrivial task that has been attracting significant interest from the computer graphics and vision communities since the first paper in 2020 \citep{mildenhall2020nerf}. We report the PSNR of the NeRF model on fern dataset (\url{https://github.com/bmild/nerf}) that was generated by the authors of \citep{mildenhall2020nerf} under the effect of PCM noise in Table~\ref{tab:nerf}. 
Though the PSNR is affected severely by the PCM noise when there is no protection, our robust strategies help to preserve the PSNR.
In particular, the PSNR is $24.75$ dB with an average number of 4.5 cells, which is even higher than the baseline (without noise) PSNR $24.73$ dB. 
Additionally, we note that we would normally expect to see a more severe effect on PSNR by the PCM noise.
This is because the exact values of the model outputs matter more in a regression setting than in a discriminative one, since the classifier's accuracy can remain unchanged even with small perturbations to the logits.
However, as can be seen from Table~\ref{tab:nerf}, PSNR is still within an acceptable range once we apply the robust strategies. We believe this is due to the discretization at the evaluation phase, which may cancel out some noise on the output.     

\setlength{\tabcolsep}{7pt}
\begin{table}[h]
\centering
\begin{tabular}{lc}
\toprule
& PSNR (dB) \\ \midrule
No protection (32 cells) & 5.66      \\ 
SP+AM+AR (4 cells) & 23.08       \\
SP+AM+AR (3 cells)       &  20.41 \\
SP+AM+AR (2 cells)       &  18.20      \\
SP+AM+AR+Sens. (4.5 cells)    &  24.75  \\
SP+AM+AR+Sens. (3.5 cells) &  22.98                  \\
SP+AM+AR+Sens. (2.5 cells) & 20.65 \\
\bottomrule
\\
\end{tabular}
\caption{NeRF \citep{mildenhall2020nerf} model on fern dataset (\url{https://github.com/bmild/nerf}). A sample of PSNR vs. average number of cells required to store 1) the continuous weight values and 2) the additional bits. Baseline PSNR (without noise) is 24.73 dB. SP: sign protection, AM: adaptive mapping, AR: sparsity-driven adaptive redundancy, Sens.: sensitivity-driven adaptive redundancy. Higher is better.} 
\label{tab:nerf}
\end{table}

As a final note, the storage density for the NeRF model or other 3D regression models could be improved further through model compression techniques such as \citep{isik2021neural, isik2021lvac, bird20213d}. 

\subsection{End-to-End Learning} Finally, we explore the effectiveness of the end-to-end learning scheme in which we train an autoencoder on a \textit{set of model weights}.
We generate two sets of 2-D Gaussian mixtures: an ``easy" task in which we first sample a two-dimensional mean vector $\mu_1 \in \mathbb{R}^2$ where $\mu_{1i} \sim \mathcal{U}[-1,0)$ and $\mu_2 \in \mathbb{R}^2$ where $\mu_{2i} \sim \mathcal{U}[0,1)$ for $i=\{1,2\}$. After sampling these means, we draw a set of 50K points for the two mixture components: $x_1 \sim \mathcal{N}(\mu_1, I)$ and $x_2 \sim \mathcal{N}(\mu_2,I)$. This ensures that the two mixture components are well-separated. For the ``hard" task, we sample overlapping means: both components of $\mu_1$ and $\mu_2$ are drawn from $\mathcal{U}[-1,+1]$ before sampling from their respective mixture components. We generate 50 datasets per task, where each dataset has 50K data points. 

We train 50 different logistic regression models on each of the datasets for both the ``easy" and ``hard" tasks (each model has 3 parameters). 
After training the logistic regression models, we use an autoencoder with MLP encoder and MLP decoder architectures and ReLU nonlinearities, as shown in Table~\ref{table:mlp_ae_arch}. The autoencoder for both the "easy" and "hard" tasks is trained for 10 epochs with a batch size of 100 using the Adam optimizer with learning rate = 0.001, $\beta_1 = 0.9$, $\beta_2 = 0.999$, and early stopping on a held-out validation set. 

\begin{figure}[!h]
    \centering
        \includegraphics[width=.7\textwidth]{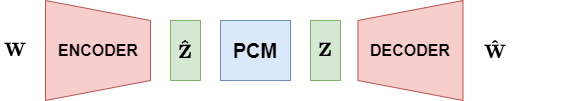}
    \caption{Illustration of the autoencoder used for the logistic regression experiments. The input weight $W$ is mapped to a compressed representation $\hat{Z}$ by an encoder module, which is then perturbed by the PCM (or analogously, Gaussian) noise channel to become a perturbed representation $Z$. This perturbed representation is then passed to the decoder, which produces a reconstructed weight $\hat{W}$.}
    \label{fig:appendix_logreg}
   \end{figure}

\begin{table}[h]
\centering
\begin{tabular}{c|c}
\hline
\textbf{Name}& \textbf{Component}\\
\hline
(Encoder) Input Layer & Linear $3 \rightarrow 1$, ReLU \\
\hline
(Encoder) Hidden Layer & Linear $1 \rightarrow 1$ \\
\hline
(Decoder) Hidden Layer & Linear $1 \rightarrow 1$, ReLU \\
\hline
(Decoder) Output Layer & Linear $1 \rightarrow 3$ \\
\hline
\end{tabular}
\caption{MLP-based autoencoder architecture for synthetic experiments, trained on 50 logistic regression classifiers per task.}
\label{table:mlp_ae_arch}
\end{table}
We test the autoencoder on both Gaussian and PCM noise: that is, we corrupt the 1-dimensional encoder output (z-representation of the classifier weights) with the appropriate perturbation before passing in the encoded representation into the decoder. Figure~\ref{fig:appendix_logreg} provides an illustration of the autoencoding process for the PCM array, which is analogous to the Gaussian noise setup. 
For the easy task with Gaussian noise, the autoencoder achieves an accuracy of 95.2\%; for PCM noise, it achieves 91.8\% accuracy. For the hard task with Gaussian noise, the autoencoder achieves an average accuracy of 78.8\% across all classifiers; for PCM channel noise, it achieves 78.6\% on average across all 50 classifiers. 

\label{appendix_logreg}

\begin{figure}[h]
    \centering
        \subfigure[Logistic regression experiment, easy task. ]{\includegraphics[width=.47\textwidth]{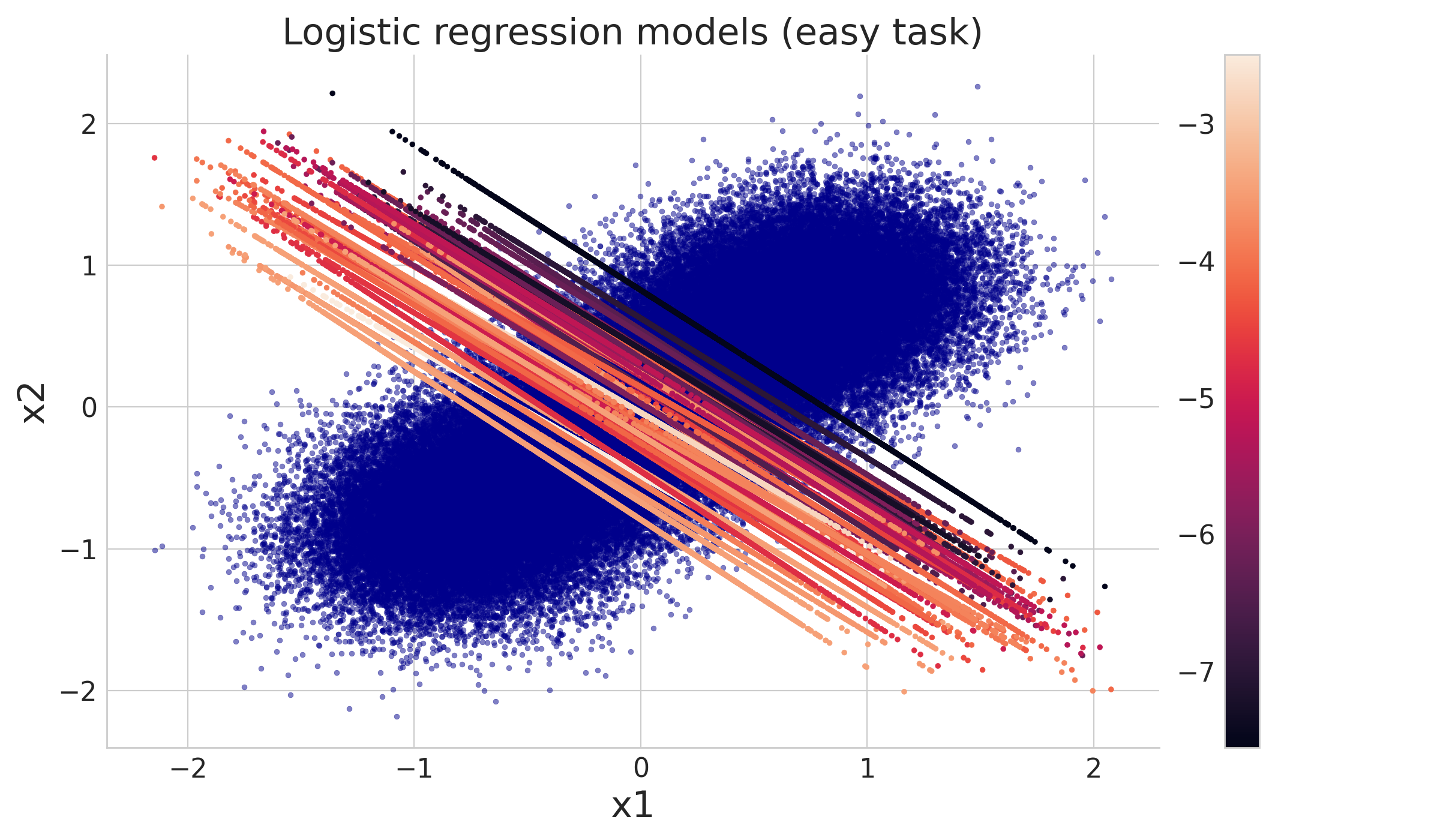}}
        \subfigure[Logistic regression experiment, hard task. ]{\includegraphics[width=.47\textwidth]{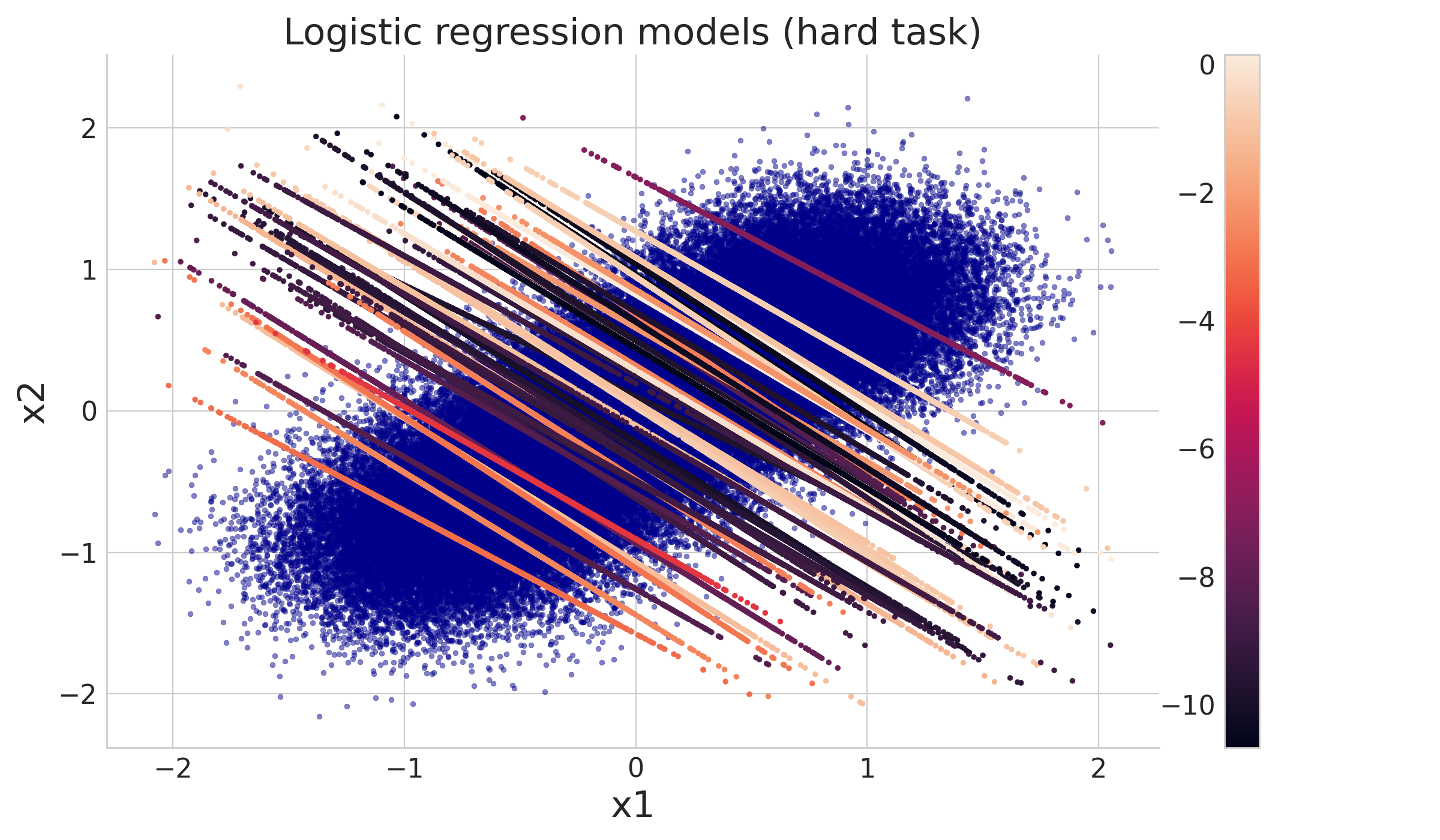}}
        \subfigure[Encoded classifier with large magnitude. ]{\includegraphics[width=.47\textwidth]{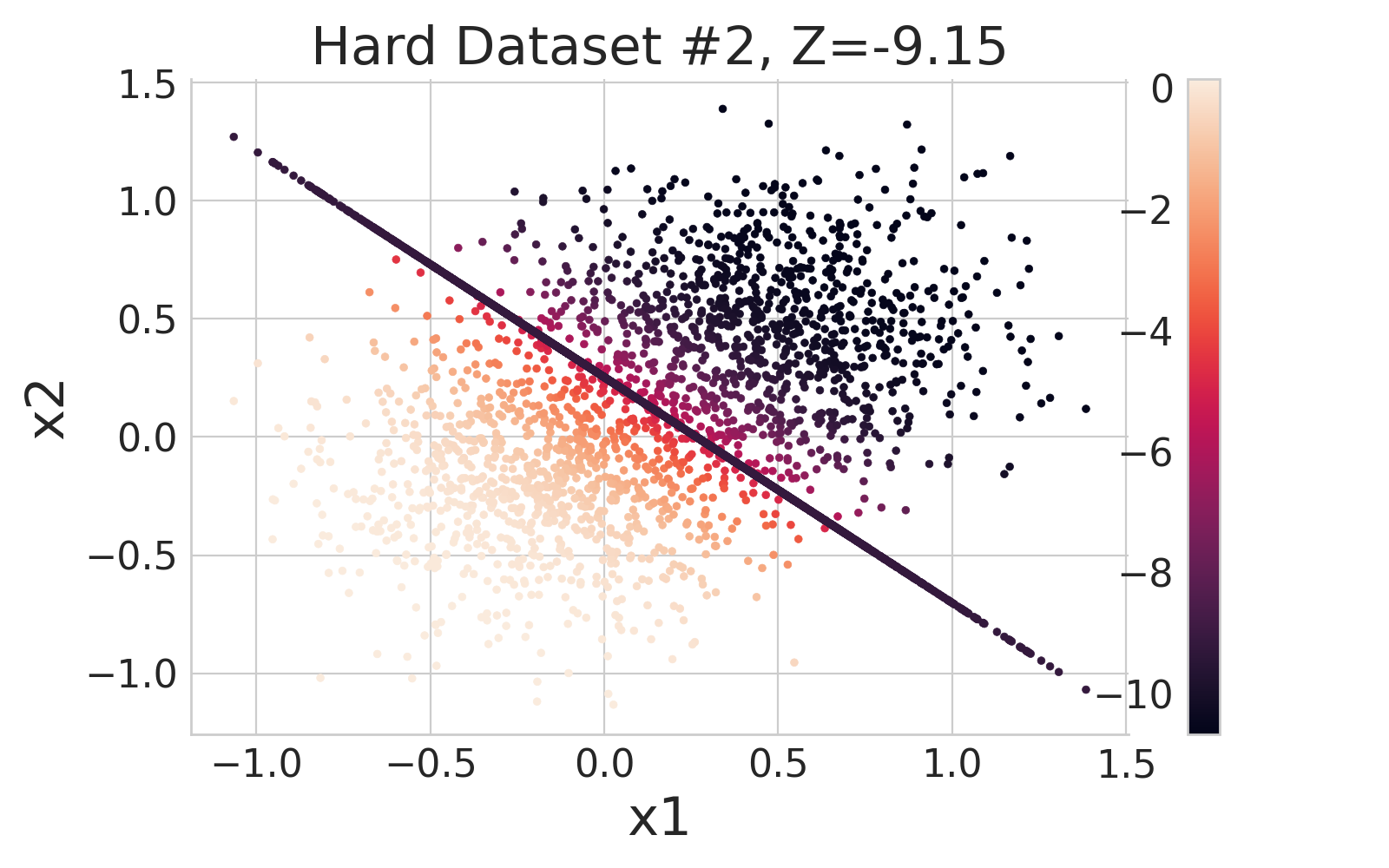}}
        \subfigure[Encoded classifier with small magnitude. ]{\includegraphics[width=.47\textwidth]{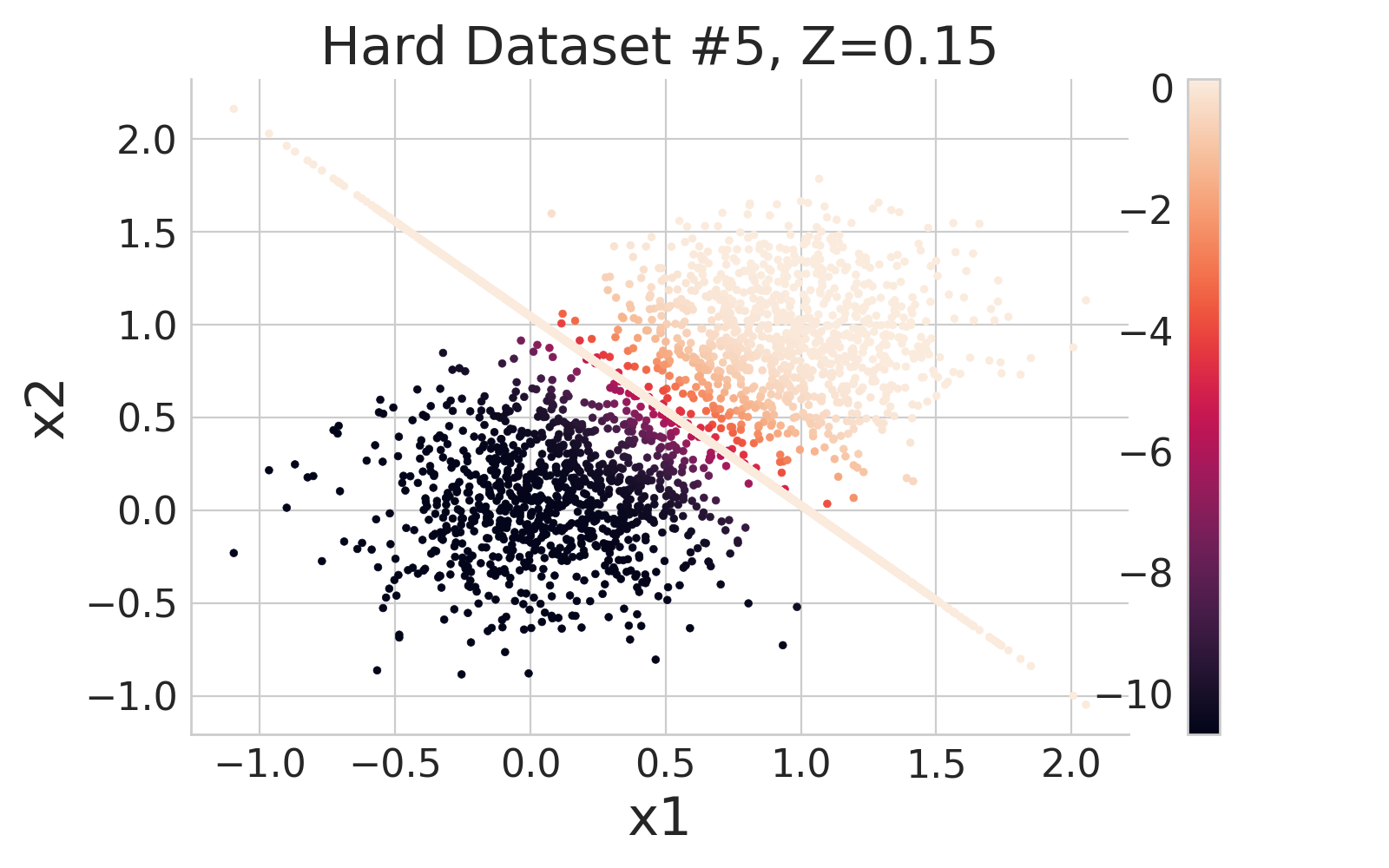}}
    \caption{\textbf{(a-b)} Qualitative visualizations of the learned representations in the logistic regression experiment. (a) shows all 50 datasets with the true decision boundaries colored by the magnitude of their z-representations for the ``easy task", while (b) shows the analogous plot for the harder task with overlapping means. \textbf{(c-d)} Qualitative visualizations of the learned representations for the hard logistic regression task. We find in (c) that classifiers with large magnitudes in z-space have the positive labels to the left of the decision boundary, while (d) those with small magnitudes have the positive labels to the right of the decision boundary.}
    \label{fig:logreg_z}
   \end{figure}

Interestingly, we find that the 1-D classifier representations in the autoencoder's latent space is semantically meaningful. In Figure~\ref{fig:logreg_z}, we qualitatively analyze the learned representations of the logistic regression classifier weights. In Figure~\ref{fig:logreg_z}(a), we plot all 50 datasets of the Gaussian mixtures (``easy task") as well as the true decision boundaries for each of the 50 logistic regression models, each boundary colored by the magnitude of its z-representation as learned by the autoencoder. We plot the same for the ``hard task" in Figure~\ref{fig:logreg_z}(b). For the easy task, we note that the classifiers are encoded by their relative location in input-space: those that are in the lower left corner of the scatter plot have smaller magnitudes than those on the upper right. For the hard task, however, the z-representations appear to be fairly random -- at a first glance, there does not appear to be a particular correlation between the magnitudes of the z-encodings and the original classifier weights. 

We further explore this phenomenon for the hard task in Figure~\ref{fig:logreg_z}(c-d). For two particular datasets (though the trend holds across all 50 datasets), we color the original data points by their mixture component as well as the true decision boundary by the magnitude of its z-encoding. As shown in Figure~\ref{fig:logreg_z}(c) and (d), we find that the autoencoder has learned to map all classifiers with the positive class to the left of the decision boundary to z-representations with large magnitude; conversely, those with the positive class to the right of the decision boundary are encoded to z-representations with smaller magnitudes. Through this preliminary investigation, we demonstrate that the end-to-end approach for learning both the compression scheme while taking the physical constraints of the storage device into account shows promise.

\section{Related Work}
\label{related}
In this section, we briefly summarize the related work in model compression and analog computing in NNs. 
\paragraph{Model Compression.}
Compression of NN parameters has a rich history starting from the early works of \citep{OBD,OBS}, with techniques such as pruning \citep{ frankle2018lottery, deep_compression,han2015learning, isik2021successive, isik2022information, snip,  woodfisher, hydra, kundu2021dnr, isik2023sparse, barnes2020rtop, pase2022efficient}, quantization, \citep{adaptiveQuant,scalableQuant, choi2020universal, QuantNoise, young2020transform} and KD \citep{distillation,polino2018model,xie2020self} among others \citep{deng2020model}. 
\cite{adversarialdistil} explores KD as a way to encourage robustness to adversarial perturbations in input space, while we specifically desire robustness to weight perturbations.
Although probabilistic approaches to model compression have been explored in \cite{louizos2017bayesian,reagan2018weightless,havasi2018minimal}, we additionally consider the physical constraints of the storage device used for memory as part of our learning scheme.
Our end-to-end approach is most similar to \cite{oktay2019scalable}, in that they also learn a decoder to map NN weights into a latent space prior to quantization.
However, our method is different in that our autoencoder also learns the appropriate compression scheme (with an encoder), we inject noise into our compressed weights rather than quantizing, and we do not require training a NN from scratch.

\paragraph{Analog Computing in NNs.} 
A complementary line of work utilizes analog components in NN training and/or inference \citep{Joshi2020AccurateDN, leblebici, onchiplearning,binas2016precise, du2018analog, AnalogMemoryAcc,  zhou2021information}.
The common theme is performing computation in the analog domain
to reduce the overall computation power, but this procedure may be noisy or inflexible.
In contrast, our work focuses on \textit{storing} NN models in analog cells --
once the parameters are read from memory, the actual computation happens in the digital domain.


\section{Discussion and Conclusion}
\label{conclusion}
In this work, we formalized the problem of jointly optimizing model compression and memory resource allocation on noisy analog storage devices.
We introduced novel coding techniques for preserving NN accuracy even in the presence of weight perturbations, and demonstrated their effectiveness on models trained on MNIST, CIFAR-10, and ImageNet. Additionally, we explored different training strategies that can be coupled with existing compression techniques such as distillation, and provided an initial foray into a fully end-to-end learning scheme for the joint problem. 

\paragraph{Limitations.} First, the actual deployment of our approach may require some level of quantization in practice -- directly writing analog values on a physical storage device requires complex read/write circuitry, which may not be feasible on current systems. For future work, we aim to investigate this bottleneck on physical hardware devices. Second, our end-to-end learned compression scheme assumes that all models share identical structures and are trained on the same dataset. Extending our framework to handle models trained on various datasets/tasks is an exciting research direction. 

\paragraph{Broader Impacts.} Although our work aims to reduce power and memory consumption through more efficient NN model compression, it is still susceptible to propagating existing biases in the original trained network.
While the research community has mainly evaluated the success of NN compression methods by only considering the compression ratio-accuracy trade-off, a recent study has shown that \emph{existing compression methods may disproportionately impact different subgroups of the data} \citep{hooker2020characterising}. 
Unfortunately, this implies that we will not be able to detect or prevent potential amplification of existing biases once the network is deployed.
This speaks to the fundamental importance in the careful curation of datasets and selection of training objectives to mitigate model bias.
\section{Acknowledgement}
The authors would like to thank TSMC Corporate Research for technical discussions; and Robert M. Radway and Pulkit Tandon for their constructive feedback. BI is supported by the Stanford Graduate Fellowship and a Meta research award. KC is supported by the NSF GRFP, Stanford Graduate Fellowship, and Two Sigma Diversity PhD Fellowship. The Stanford authors’ work was supported by NSF ($\#$1651565, $\#$1522054, $\#$1733686), ONR (N00014-19-1-2145), AFOSR (FA9550-19-1-0024), ARO (W911NF2110125), and Amazon AWS.

\bibliography{icml}
\bibliographystyle{icml2021}
\clearpage
\section*{Appendix}
\label{appendix}
\renewcommand{\thesubsection}{\Alph{subsection}}


\subsection{Additional Experimental Details}
We conducted our experiments on NVIDIA Titan XP GPUs on an internal cluster server. We used 2 GPUs for ImageNet experiments and 1 GPU for the rest of the experiments.
In the following subsection, we provide additional details on the models, model architectures, and hyperparameters used in our experiments. 
\label{hyperparams}
\subsubsection{MNIST} 
The architectures for LeNet and the MLP are shown in Tables~\ref{table:lenet_arch} and~\ref{table:mlp_arch} respectively:

\begin{table}[h!]
\centering
\begin{tabular}{c|c}
\hline
\textbf{Name}& \textbf{Component}\\
\hline
conv1 & [$5 \times 5$ conv, 20 filters, stride 1], ReLU, $2 \times 2$ max pool \\
\hline
conv2 & [$5 \times 5$ conv, 50 filters, stride 1], ReLU, $2 \times 2$ max pool \\
\hline
Linear & Linear $800 \rightarrow 500$, ReLU \\
\hline
Output Layer & Linear $500 \rightarrow 10$ \\
\hline
\end{tabular}
\caption{LeNet architecture for MNIST experiments.}
\label{table:lenet_arch}
\end{table}

\begin{table}[h!]
\centering
\begin{tabular}{c|c}
\hline
\textbf{Name}& \textbf{Component}\\
\hline
Input Layer & Linear $784 \rightarrow 100$, ReLU \\
\hline
Hidden Layer & Linear $100 \rightarrow 100$, ReLU \\
\hline
Output Layer & Linear $100 \rightarrow 10$ \\
\hline
\end{tabular}
\caption{MLP architecture for MNIST experiments.}
\label{table:mlp_arch}
\end{table}
For both the LeNet and MLP classifiers, we use a batch size of 100 and train for 100 epochs, early stopping at the best accuracy on the validation set. We use the Adam optimizer with learning rate $=0.001$, and $\beta_1=0.9, \beta_2=0.999$ with weight decay $=5e^{-4}$. For knowledge distillation, we use a temperature parameter of $T = 1.5$ and equally weight the contributions of the student network's cross entropy loss and the distillation loss ($\lambda=0.5$). 

\subsubsection{CIFAR-10} We provide the architectural details for the ResNet-18 and the slim ResNet-20 used in our experiments in Tables~\ref{table:resnet_arch} and~\ref{table:resnet_arch2} below:
\begin{table}[h!]
\centering
\begin{tabular}{c|c}
\hline
\textbf{Name}& \textbf{Component}\\
\hline
conv1 & $3\times3$ conv, 64 filters. stride 1, BatchNorm \\
\hline
Residual Block 1 & 
$
\begin{bmatrix}
    3 \times 3 \text{ conv, } 64 \text{ filters} \\
    3 \times 3 \text{ conv, } 64 \text{ filters}
\end{bmatrix}
\times 2$ \\
\hline
Residual Block 2 & 
$
\begin{bmatrix}
    3 \times 3 \text{ conv, } 128 \text{ filters} \\
    3 \times 3 \text{ conv, } 128 \text{ filters}
\end{bmatrix}
\times 2$ \\
\hline
Residual Block 3 & $
\begin{bmatrix}
    3 \times 3 \text{ conv, } 256 \text{ filters} \\
    3 \times 3 \text{ conv, } 256 \text{ filters}
\end{bmatrix}
\times 2$ \\
\hline
Residual Block 4 & $
\begin{bmatrix}
    3 \times 3 \text{ conv, } 512 \text{ filters} \\
    3 \times 3 \text{ conv, } 512 \text{ filters}
\end{bmatrix}
\times 2$ \\
\hline
Output Layer & $4 \times 4$ average pool stride 1, fully-connected, softmax \\
\hline
\end{tabular}
\caption{ResNet-18 architecture for CIFAR-10 experiments.}
\label{table:resnet_arch}
\end{table}

\begin{table}[h!]
\centering
\begin{tabular}{c|c}
\hline
\textbf{Name}& \textbf{Component}\\
\hline
conv1 & $3\times3$ conv, 16 filters. stride 1, BatchNorm \\
\hline
Residual Block 1 & 
$
\begin{bmatrix}
    3 \times 3 \text{ conv, } 16 \text{ filters} \\
    3 \times 3 \text{ conv, } 16 \text{ filters}
\end{bmatrix}
\times 2$ \\
Residual Block 2 & 
$
\begin{bmatrix}
    3 \times 3 \text{ conv, } 32 \text{ filters} \\
    3 \times 3 \text{ conv, } 32 \text{ filters}
\end{bmatrix}
\times 2$ \\
\hline
Residual Block 3 & $
\begin{bmatrix}
    3 \times 3 \text{ conv, } 64 \text{ filters} \\
    3 \times 3 \text{ conv, } 64 \text{ filters}
\end{bmatrix}
\times 2$ \\
\hline
Output Layer & $7 \times 7$ average pool stride 1, fully-connected, softmax \\
\hline
\end{tabular}
\caption{Slim ResNet-20 architecture for CIFAR-10 experiments.}
\label{table:resnet_arch2}
\end{table}
For both ResNet-18 and slim ResNet-20, we use a batch size of 128 and train for 350 epochs, early stopping at the best accuracy on the validation set. We use SGD with learning rate $=0.1$, and momentum $=0.9$ and weight decay $=5e^{-4}$. 

\subsubsection{ImageNet}
We provide the architectural details for the ResNet-50 used in our experiments in Table~\ref{table:resnet_arch3}.

\begin{table}[h!]
\centering
\begin{tabular}{c|c}
\hline
\textbf{Name}& \textbf{Component}\\
\hline
conv1 & $3\times3$ conv, 64 filters. stride 1, BatchNorm \\
\hline
Residual Block 1 & 
$
\begin{bmatrix}
    1 \times 1 \text{ conv, } 64 \text{ filters} \\
    3 \times 3 \text{ conv, } 64 \text{ filters} \\
    1 \times 1 \text{ conv, } 256 \text{ filters}
\end{bmatrix}
\times 3$ \\
\hline
Residual Block 2 & 
$
\begin{bmatrix}
    1 \times 1 \text{ conv, } 128 \text{ filters} \\
    3 \times 3 \text{ conv, } 128 \text{ filters} \\
    1 \times 1 \text{ conv, } 512 \text{ filters}
\end{bmatrix}
\times 4$ \\
\hline
Residual Block 3 & $
\begin{bmatrix}
    1 \times 1 \text{ conv, } 256 \text{ filters} \\
    3 \times 3 \text{ conv, } 256 \text{ filters} \\
    1 \times 1 \text{ conv, } 1024 \text{ filters}
\end{bmatrix}
\times 6$ \\
\hline
Residual Block 4 & $
\begin{bmatrix}
    1 \times 1 \text{ conv, } 512 \text{ filters} \\
    3 \times 3 \text{ conv, } 512 \text{ filters} \\
    1 \times 1 \text{ conv, } 2048 \text{ filters}
\end{bmatrix}
\times 3$ \\
\hline
Output Layer & $4 \times 4$ average pool stride 1, fully-connected, softmax \\
\hline
\end{tabular}
\caption{ResNet-50 architecture for ImageNet experiments.}
\label{table:resnet_arch3}
\end{table}
We use the pretrained ResNet-50 from PyTorch (\texttt{https://github.com/pytorch/vision/blob/\\master/torchvision/models/resnet.py}), with a batch size of 64. For the robust training experiments, we retrain the model for 20 epochs with early stopping at best accuracy. We use SGD with learning rate $=0.001$, and momentum $=0.9$ and weight decay $=5e^{-4}$. 

\subsection{Additional Experimental Results}
\label{confidence}

\subsubsection{Sparsity- and Sensitivity-Driven Protection}
    \begin{figure*}[h]
        \centering %
        \subfigure[Perturbation by PCM cells.]{\includegraphics[width=.47\textwidth]{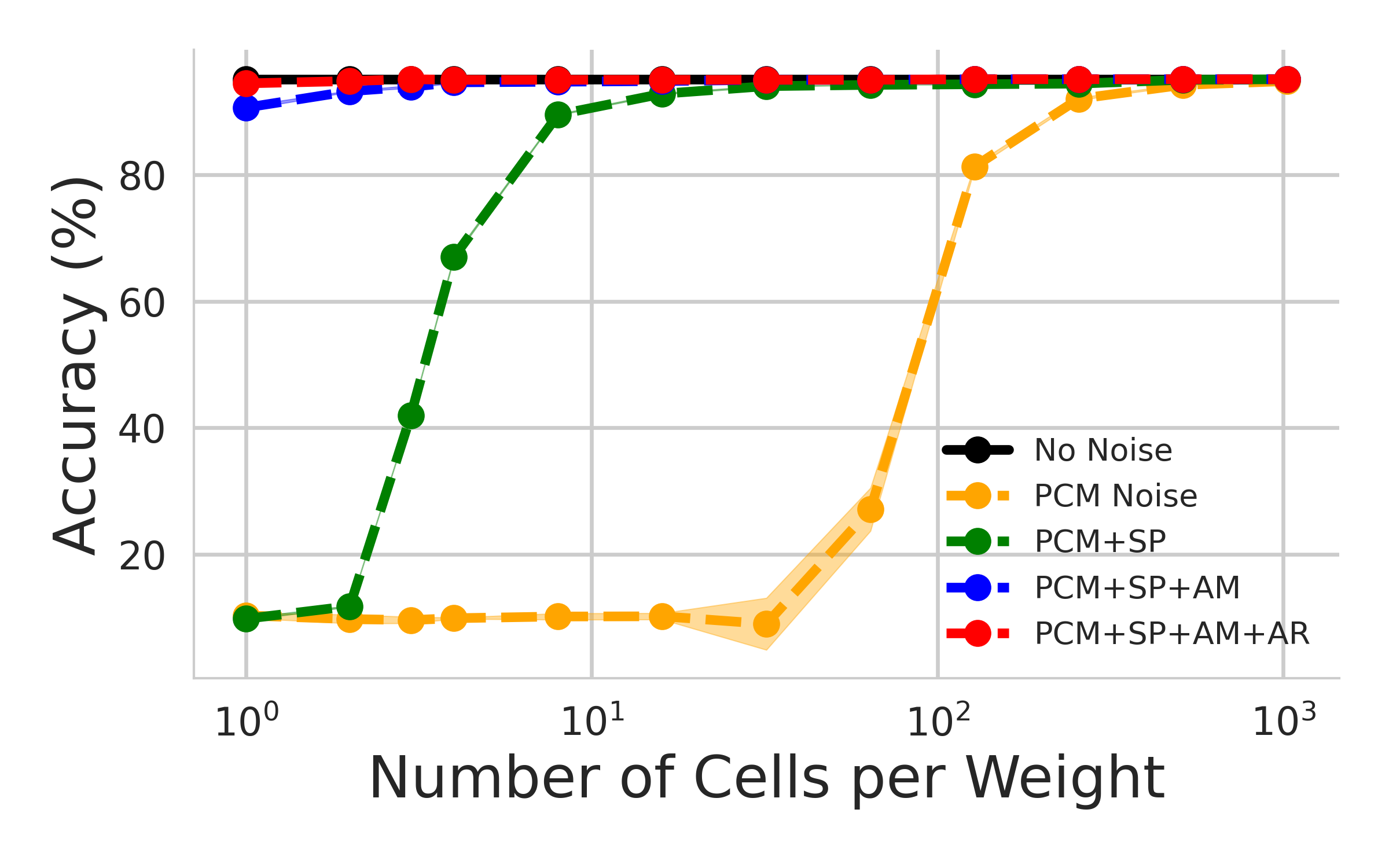}}
        \subfigure[Perturbation by Gaussian noise.]{\includegraphics[width=.47\textwidth]{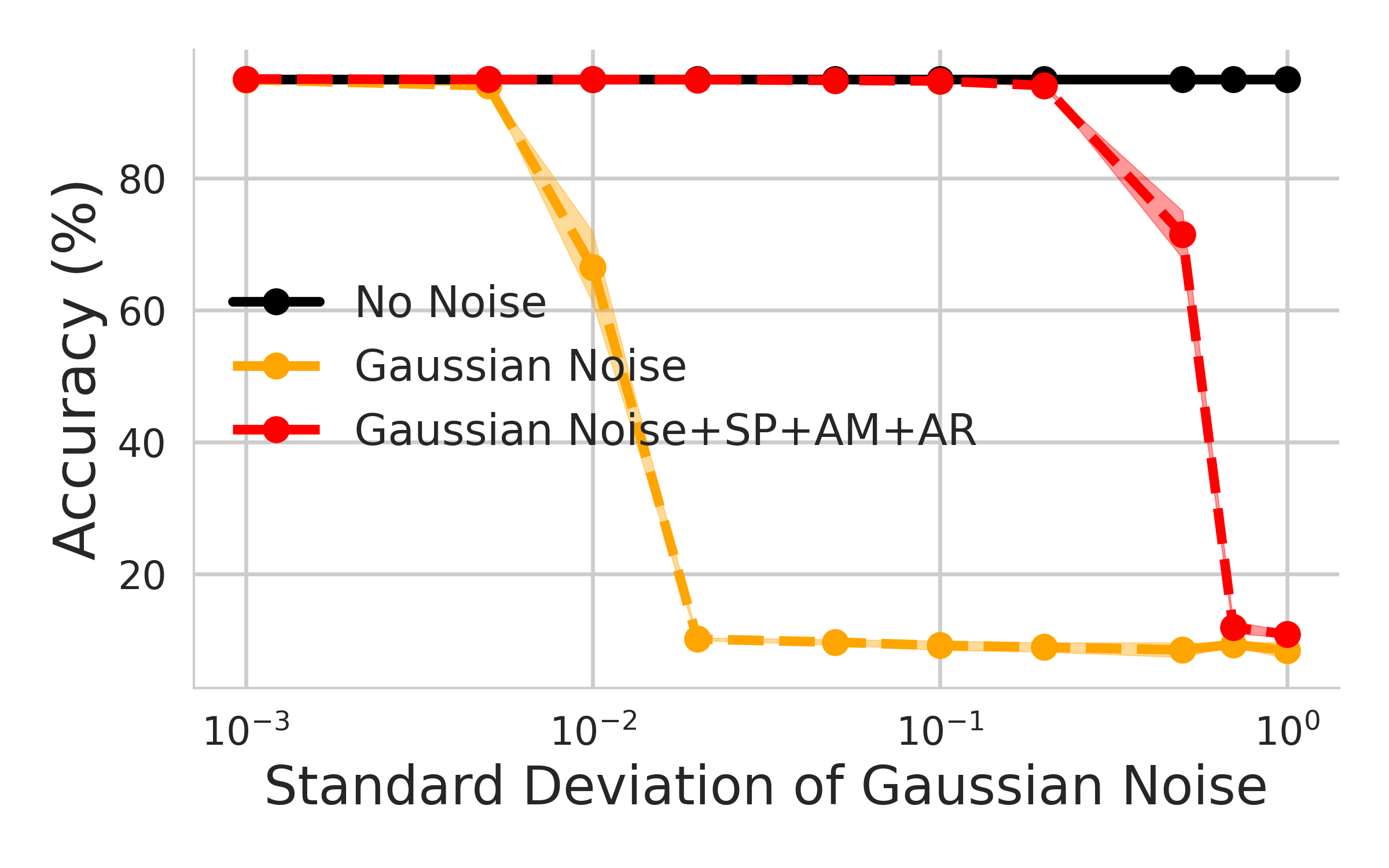}}
    \caption{Accuracy of ResNet-18 on CIFAR-10 when weights are perturbed by (a) PCM cells, (b) Gaussian noise. SP: sign protection, AM: adaptive mapping, AR: sparsity-driven adaptive redundancy. Experiments are conducted three times. 
    }\label{fig:comparison_cifar_app}
   \end{figure*}
   
\setlength{\tabcolsep}{3pt}
\begin{table*}[t]
\centering
\resizebox{\textwidth}{!}{
\begin{tabular}{lcccccccccc}
\toprule
                            & 512 cells      & 64 cells    & 32 cells    & 16 cells         & 8 cells     & 4 cells   & 3 cells   & 2 cells & 1 cell & Additional Bits\\ \midrule
No Protection                   & 94.2($\pm$ 0.1)& 27.1 ($\pm$ 3.4) & 9.0 ($\pm$ 4.1)       & 10.2 ($\pm$ 0.5)  & 10.2 ($\pm$ 0.5)  & 9.9 ($\pm$ 0.1)    & 9.6 ($\pm$ 0.5) & 9.8 ($\pm$ 0.7) & 10.3 ($\pm$ 0.4) & 0\\
SP                      & 95.0 ($\pm$ 0.0) & 94.2 ($\pm$ 0.1)      & 94.00 ($\pm$ 0.2)       & 92.80 ($\pm$ 0.3)  & 89.50 ($\pm$ 0.6)  & 67.00 ($\pm$ 0.2)    & 41.90 ($\pm$ 0.5) & 11.80 ($\pm$ 5.1) & 9.80 ($\pm$ 1.5) & 1\\ 
AM+AR                   & 95.0 ($\pm$ 0.0) & 94.8 ($\pm$ 0.1)       & 94.70 ($\pm$ 0.1)       & 94.40 ($\pm$ 0.1)  & 93.70 ($\pm$ 0.2)  & 93.10 ($\pm$ 0.2)    & 92.70 ($\pm$ 0.2) & 89.20 ($\pm$ 2.6) & 58.00 ($\pm$ 8.2) & 1\\
SP+AM                   & 95.1 ($\pm$ 0.0) & 95.0 ($\pm$ 0.0)                        & 95.00 ($\pm$ 0.1)       & 94.80 ($\pm$ 0.1)  & 94.70 ($\pm$ 0.1)  & 94.60 ($\pm$ 0.1)    & 93.90 ($\pm$ 0.2) & 93.20 ($\pm$ 0.1) & 90.60 ($\pm$ 0.3) & 2\\ 
SP+AM+AR                & 95.1 ($\pm$ 0.0)& 95.0 ($\pm$ 0.0)                       & \textbf{95.00} ($\pm$ 0.1)       & \textbf{95.00} ($\pm$ 0.1)  & \textbf{95.00} ($\pm$ 0.1)  & \textbf{95.00} ($\pm$ 0.1)    & \textbf{95.10} ($\pm$ 0.1) & \textbf{94.80} ($\pm$ 0.1) & \textbf{94.44} ($\pm$ 0.2) & 2\\

SP+AM+AR+Sens.    & 95.1 ($\pm$ 0.0) & 95.1 ($\pm$ 0.0)                       & \textbf{95.10} ($\pm$ 0.1)       & \textbf{95.07} ($\pm$ 0.1)  & \textbf{95.11} ($\pm$ 0.1)  & \textbf{95.03} ($\pm$ 0.2)    & \textbf{95.14} ($\pm$ 0.1) & \textbf{94.80} ($\pm$ 0.1) & \textbf{94.95} ($\pm$ 0.3) & 3\\
\bottomrule
\\
\end{tabular}
}
\caption{Accuracy of ResNet-18 on CIFAR-10 when weights are perturbed by the PCM cells. Baseline accuracy (when there is no noise) is $95.10 \%$. SP: sign protection, AM: adaptive mapping, AR: sparsity-driven adaptive redundancy, Sens.: sensitivity-driven adaptive redundancy. Reported results are averaged over three experimental runs.}
\label{tab:cifar_sparsity_app}
\end{table*}

\setlength{\tabcolsep}{3pt}
\begin{table*}[!t]
\centering
\resizebox{\textwidth}{!}{
\begin{tabular}{lccccccccccc}
\toprule
                           & 512 cells           & 64 cells                  & 32 cells                & 16 cells              & 8 cells          & 4 cells               & 3 cells           &2 cells                & 1 cell & Additional Bits\\ \midrule
No Protection              & 0.1($\pm$ 0.0)& 0.1 ($\pm$ 0.0)            & 0.1 ($\pm$ 0.0)       & 0.1 ($\pm$ 0.0)  & 0.1 ($\pm$ 0.0)  & 0.1 ($\pm$ 0.0)     & 0.1 ($\pm$ 0.0) & 0.1 ($\pm$ 0.0) & 0.1 ($\pm$ 0.0) & 0\\
SP                         & 4.7($\pm$ 0.0) & 0.9  ($\pm$ 0.0)          & 0.1 ($\pm$ 0.0)       & 0.1 ($\pm$ 0.0)  & 0.1 ($\pm$ 0.0)  & 0.1 ($\pm$ 0.0)     & 0.1 ($\pm$ 0.0) & 0.1 ($\pm$ 0.0) & 0.1 ($\pm$ 0.0) & 1\\ 
AM+AR                      & 75.0 ($\pm$ 0.5) & 66.7  ($\pm$ 0.8)         & 66.4 ($\pm$ 2.5)       & 63.2 ($\pm$ 2.5)  & 57.5 ($\pm$ 2.2)  & 34.1 ($\pm$ 3.4)     & 22.4 ($\pm$ 3.0) & 3.7 ($\pm$ 0.0) & 0.1 ($\pm$ 0.0) & 2\\
SP+AM                      & 70.6 ($\pm$ 0.7) & 69.4  ($\pm$ 1.1)         & 55.1 ($\pm$ 3.2)       & 51.3 ($\pm$ 2.4)  & 34.6 ($\pm$ 1.8)  & 28.0 ($\pm$ 5.6)     & 17.6 ($\pm$ 4.1) & 4.6 ($\pm$ 0.0) & 0.2 ($\pm$ 0.0) & 2\\
SP+AM+AR                   & 75.4 ($\pm$ 0.2) & 74.5 ($\pm$ 0.1)          & 74.2 ($\pm$ 0.4)       & 73.8 ($\pm$ 0.0)  & 73.8 ($\pm$ 0.2)  & 72.8 ($\pm$ 0.0)     & 72.2 ($\pm$ 0.4) & 70.2 ($\pm$ 0.0) & 66.0 ($\pm$ 0.8) & 1\\
SP+AM+AR+Sens.    & \textbf{75.8} ($\pm$ 0.0) & \textbf{75.6}    ($\pm$ 0.0)                    & \textbf{75.3} ($\pm$ 0.0)       & \textbf{74.8} ($\pm$ 0.1)  & \textbf{74.9} ($\pm$ 0.1)  & \textbf{74.5} ($\pm$ 0.1)    & \textbf{73.7} ($\pm$ 0.0) & \textbf{73.0} ($\pm$ 0.2) & \textbf{68.8} ($\pm$ 0.3) & 3\\
\bottomrule
\\
\end{tabular}
}
\caption{Accuracy of ResNet-50 on ImageNet when weights are perturbed by the PCM cells. Baseline accuracy (when there is no noise) is $76.6 \%$. SP: sign protection, AM: adaptive mapping, AR: sparsity-driven adaptive redundancy, Sens.: sensitivity-driven adaptive redundancy. Reported results are averaged over three experimental runs.}
\label{tab:imagenet_sparsity_app}
\end{table*}

We give the full results of sparsity-driven and sensitivity-driven protection experiments on CIFAR-10 and ImageNet with confidence intervals in Table~\ref{tab:cifar_sparsity_app} and Table~\ref{tab:imagenet_sparsity_app}. In Figure~\ref{fig:comparison_cifar_app}, we present experimental results with ResNet-18 on CIFAR-10. As can be seen from Figure~\ref{fig:comparison_cifar_app}(a), our strategies, namely sign protection, adaptive mapping and adaptive redundancy, reduce the number of PCM cells per weight required to preserve accuracy from $1024$ to $1$. We also test our strategies against Gaussian noise. In the Gaussian experiments, we consider hypothetical storage devices with white Gaussian noise where the standard deviation of the overall noise on the output ($\hat{\epsilon}(r, \alpha)$ ``not input-dependent'') can be reduced by using the channel multiple times (Method \#1) and by using the channel in the its power limit (Method \#2). Figure~\ref{fig:comparison_cifar_app}(b) shows that our strategies increase the standard deviation threshold, where the NN performance sharply drops, from $0.005$ to $0.2$, i.e., robustness increases by $40$ times. We note that we present results of Gaussian noise experiments to be comparable to future work although this scenario is not realistic.  


\subsubsection{Robust Pruning}
\label{pruning_app}
We give the results of robust pruning experiment in Table~\ref{tab:cifar_pruning_app}. We apply one-shot pruning followed by $20$ epochs of retraining. When sparsity- and sensitivity-driven strategies are applied, $2.05$ cells per weight are enough to preserve the original accuracy of the pruned model ($94.8 \%$). Moreover, if we use only $1.35$ cells per weight on average, the accuracy drop is only $0.1 \%$, which is insignificant considering the fact that pruning reduces the accuracy by $0.3\%$ (from $95.1\%$ to $94.8\%$). We note that this performance degradation due to pruning can be eliminated with robust training, explained in Section~\ref{robust_train}. With this strategy, we can reduce the number of cells required to store each weight from 16 (in digital storage) to $1.35$ with analog storage combined with our robust strategies and pruning. This corresponds to a $11.85$ times more efficient storage.  


\setlength{\tabcolsep}{3pt}
\begin{table*}[!t]
\centering
\resizebox{0.6\textwidth}{!}{
\begin{tabular}{lcccccc}
\toprule
&                           2.05 cells                    & 1.55 cells                        & 1.95 cells               & 1.45 cells           & 1.85 cells      & 1.35 cells       \\ \midrule
SP+AM+AR                    &  -                         & 94.3                           & -                              & 94.5                & -             & 94.2  \\
SP+AM+AR+Sens.              &  94.8                      &   -                            & 94.6                           & -                   & 94.7           & - \\
\bottomrule
\\
\end{tabular}}
\caption{Accuracy of $90 \%$ pruned ResNet-18 on CIFAR-10 when weights are perturbed by the PCM cells vs the average number of cells required to store 1) the continues weight values of the non-pruned parameters and 2) the pruning mask. Baseline accuracy after the pruning (when there is no noise) is $94.8 \%$. SP: sign protection, AM: adaptive mapping, AR: sparsity-driven adaptive redundancy, Sens.: sensitivity-driven adaptive redundancy. 
}
\label{tab:cifar_pruning_app}
\end{table*}


\subsubsection{Robust Training}
\setlength{\tabcolsep}{3pt}
\begin{table}[!h]
\centering
\resizebox{0.8\textwidth}{!}{
\begin{tabular}{lcccc}
\toprule
                            & Naive Training       & Robust Training                        & Robust Training      & Additional\\ 
                            &(no noise)             &  (with N(0, 0.01))                    & (with N(0, 0.006))   &  Bits \\ \midrule
No Noise                    & 95.10                 & 95.50                                 & \textbf{95.60}                    &0\\
PCM Noise (No Protection)   & 9.70($\pm$ 1.0)      & 8.30 ($\pm$ 0.4)                    & 9.90 ($\pm$ 1.1)                    &1\\
PCM Noise+SP                & 9.70($\pm$ 1.5)      & 10.63 ($\pm$ 0.2)                   & 10.33 ($\pm$ 0.2)                    &2\\
PCM Noise+SP+AP             & 90.60($\pm$ 0.2)     & 94.73 ($\pm$ 0.1)                   & \textbf{94.80} ($\pm$ 0.2)                    &2\\
PCM Noise+AP                & 27.69($\pm$ 8.2)     & \textbf{86.20} ($\pm$ 0.4)           & 81.83 ($\pm$ 4.2)                    &1\\
PCM Noise+AP+Sens           & 36.21($\pm$ 3.1)     & \textbf{86.73} ($\pm$ 2.2)           & 78.93 ($\pm$ 4.4)                    &2\\
PCM Noise+SP+AP+Sens        & 94.95($\pm$ 0.3)     & \textbf{95.03} ($\pm$ 0.1)           & \textbf{95.03} ($\pm$ 0.1)                    &3\\
\bottomrule
\\
\end{tabular}
}
\caption{Accuracy of ResNet-18 on CIFAR-10 when weights are perturbed by the PCM cells. Baseline accuracy (when there is no noise at train or test time) is $95.10 \%$. SP: sign protection, AP: adaptive protection (adaptive mapping+sparsity-driven adaptive redundancy), Sens.: sensitivity-driven adaptive redundancy. Experiments are conducted three times.}
\label{tab:cifar_robust_train_app}
\end{table}

\setlength{\tabcolsep}{3pt}
\begin{table*}[!h]
\centering
\resizebox{0.8\textwidth}{!}{
\begin{tabular}{lcccc}
\toprule
                            & Naive Training       & Robust Training                        & Robust Training      & Additional \\ 
                            &(no noise)             &  (with N(0, 0.01))                    & (with N(0, 0.6))   &  Bits \\ \midrule
No Noise                    & 76.60   & 76.60                  & 76.60                    &0\\
PCM Noise (No Protection)   & 0.1($\pm$ 0.0)      & 0.1 ($\pm$ 0.0)                    & 0.1 ($\pm$ 0.0)                    &0\\
PCM Noise+SP                & 0.1($\pm$ 0.0)     & \textbf{0.4} ($\pm$ 0.0)                   & 0.2 ($\pm$ 0.0)                    &1\\
PCM Noise+AP                & 0.1($\pm$ 0.0)     & \textbf{0.2} ($\pm$ 0.0)                   & 0.1 ($\pm$ 0.0)                    &1\\
PCM Noise+SP+AP             & 66.00($\pm$ 0.8)     & 69.00 ($\pm$ 0.3)           & \textbf{69.20} ($\pm$ 0.2)                    &2\\
PCM Noise+AP+Sens           & 0.3($\pm$ 0.0)     & \textbf{0.6} ($\pm$ 0.0)                   & 0.5 ($\pm$ 0.0)                    &2\\
PCM Noise+SP+AP+Sens        & 68.80 ($\pm$ 0.3)     & \textbf{70.20} ($\pm$ 0.0)                   & \textbf{70.40} ($\pm$ 0.1)                    &3\\

\bottomrule
\\
\end{tabular}
}
\caption{Accuracy of ResNet-50 on ImageNet when weights are perturbed by the PCM cells. Baseline accuracy (when there is no noise at train or test time) is $74.4 \%$. SP: sign protection, AP: adaptive protection (adaptive mapping+sparsity-driven adaptive redundancy), Sens.: sensitivity-driven adaptive redundancy. Experiments are conducted three times.}
\label{tab:imagenet_robust_train_app}
\end{table*}

We give the full results of robust training experiments on CIFAR-10 and ImageNet with confidence intervals in Table~\ref{tab:cifar_robust_train_app} and Table~\ref{tab:imagenet_robust_train_app}. We use $\lambda=0.5$ as the coefficient of the KL regularization term in the loss function in Section~\ref{robust_train}. The level of the noise (standard deviation) injected during the training is adjusted according to $r$ and $\alpha$ values to be used at storage time since the noise at storage time has a standard deviation of  $\frac{\sigma(w_{in})}{\alpha \sqrt{r}}$. It is seen from Table~\ref{tab:cifar_robust_train_app} and Table~\ref{tab:imagenet_robust_train_app} that robust training improves the robustness of the network against noise. We have also observed that the pruned network reaches higher accuracy after robust training compared to naive training without noise injection (an increase from $90.2\%$ to $95.0\%$ without retraining). 

\subsubsection{Robust Distillation}
\label{distill_exp_app}
Knowledge distillation (KD) is a well-established NN compression method where a large teacher network is trained with $\mathcal{L}_t$ loss given by
\begin{align*}
    \mathcal{L}_t = \E_{x,y \sim p_{data}}[-\log p_{w_t}(y|x)]
\end{align*}
where $w_t$ is the weights of the teacher network, and probability of each class $i$ is the output of the high temperature softmax activation applied to logits:
\begin{align*}
    y_{i} = \frac{\exp(z_i/T)}{\sum_j \exp(z_j/T)}.
\end{align*}
where $z_i$ is the logit for class $i$. Temperature $T>1$ helps the output probabilities of the teacher network be softer. Using the same temperature, a smaller student network can be trained with the following student loss function $\mathcal{L}_s$:
\begin{align*}
    \mathcal{L}_s &= (1-\lambda)\E_{x,y \sim p_{data}}[-\log p_{w_s}(y|x)]  + \lambda \E_x[D_{KL}(p_{w_t}(y|x)||p_{w_s}(y|x))]
\end{align*}

where $w_s$ is the weights of the student network. It has been shown that distilled student network achieves a test accuracy that a teacher network with the same architecture cannot achieve. In other words, a student network distilled from a teacher network performs comparable to a larger network. This suggests that knowledge distillation can be regarded as a promising compression method. In addition to compression, distillation has been shown to be an effective method for other desired NN attributes such as generalizability and adversarial robustness \citep{adversarialdistil}. In this work, we define ``robustness" as preserving a network's downstream classification accuracy when noise is added to the weights.
This is achieved in part by robust training in the previous section where a trained network is robust to pruning and noise on the weights. Here, we present a student loss function that would make a (compressed) student network more robust to noise with no change in the teacher network training: 
\begin{align*}
    \mathcal{L}_s &= (1-\lambda)\E_{x,y \sim p_{data}}[-\log p_{g(\hat{w}_s)}(y|x)]  + \lambda \E_x[D_{KL}(p_{w_t}(y|x)||p_{g(\hat{w}_s)}(y|x))]
\end{align*}

In the experiments, we use a temperature parameter of $T = 1.5$ and equally weight the contributions of the student network's cross entropy loss and the KL term ($\lambda=0.5$). Similar to robust training, the noise level (standard deviation) during training is adjusted according to $r$ and $\alpha$ values to be used at storage time. We give the full results on CIFAR-10 with confidence intervals in Tables~\ref{tab:cifar_distill_PCM_app} and \ref{tab:cifar_distill_Gaussian_app} with weights perturbed by PCM cells and Gaussian noise, respectively. We also present the same set of experiments on MLP distilled from LeNet (on MNIST) in Tables~\ref{tab:mnist_distill_app} and \ref{tab:mnist_distill_Gaussian_app}. As in CIFAR-10 experiments, noise injection during distillation makes the student network more robust to both PCM and Gaussian noise in MNIST experiments.

\setlength{\tabcolsep}{3pt}
\begin{table*}[!t]
\centering
\resizebox{0.9\textwidth}{!}{
\begin{tabular}{ccccccc}
\toprule
                            & Number of  & Teacher                       & Teacher                               & Student                       & Noisy Student        & Add.    \\ 
                            & PCM cells   & ResNet-18                     &  ResNet-20                    & ResNet-20            & ResNet-20    & Bits                     \\ \midrule
No Noise                    & 16 & 95.70                         & 92.50                                  & 92.90                        & \textbf{93.00}     &0\\ \midrule
PCM+AP               &3   & 16.23 ($\pm$ 1.6)     & 48.38  ($\pm$ 14.3)            & 73.38     ($\pm$ 2.4)     & \textbf{81.75 } ($\pm$ 1.5)         &1\\ \midrule
\centered{PCM+SP+AP} & \centered{1 \\ 3} & \centered{93.35 ($\pm$ 0.5)  \\ 94.78 ($\pm$ 0.2)} & \centered{86.30 ($\pm$ 1.5)\\ 89.73 ($\pm$ 0.2)  } & \centered{88.58  ($\pm$ 0.4) \\ 89.98 ($\pm$ 0.3)     } & \centered{\textbf{90.65}  ($\pm$ 0.7)  \\ \textbf{91.33} ($\pm$ 0.2) } & \centered{2} \\ \midrule
PCM+AP+Sens.           & 1 & 9.60($\pm$ 0.4)      & 29.68 ($\pm$ 0.4)         & 38.18   ($\pm$ 7.2)                 & \textbf{69.49 } ($\pm$ 3.7)        &2 \\ \midrule

\centered{PCM+SP+AP+Sens.} & \centered{1 \\ 3} & \centered{93.36 ($\pm$ 0.2)\\ 94.90 ($\pm$ 0.2)} & \centered{88.40 ($\pm$ 1.4)  \\ 89.92 ($\pm$ 0.5) } & \centered{88.96 ($\pm$ 0.7)   \\ 90.44 ($\pm$ 0.6)    } & \centered{\textbf{91.10}  ($\pm$ 0.3) \\ \textbf{91.78 } ($\pm$ 0.2)} & \centered{3} \\
\bottomrule
\\
\end{tabular}
}
\caption{Accuracy of ResNet-20 distilled from ResNet-18 on CIFAR-10 when weights are perturbed by PCM. SP: sign protection, AP: adaptive protection (adaptive mapping+sparsity-driven adaptive redundancy), Sens.: sensitivity-driven adaptive redundancy. During the distillation of noisy student, a Gaussian noise with N(0,0.01) is injected onto the weights. Experiments are conducted five times.}
\label{tab:cifar_distill_PCM_app}
\end{table*}

\setlength{\tabcolsep}{3pt}
\begin{table*}[!t]
\centering
\resizebox{0.8\textwidth}{!}{
\begin{tabular}{lccccc}
\toprule
                            & Teacher                       & Teacher                               & Student                       & Noisy Student       & Additional  \\ 
                            & ResNet-18                     &  ResNet-20                    & ResNet-20            & ResNet-18                &Bits          \\ \midrule
No Noise                    & 95.70                         & 92.50                                  & 92.90                        & \textbf{93.00}     &0 \\
N(0,0.01)+No Protection     & 82.90 ($\pm$ 2.3)    &  86.44 ($\pm$ 1.6)            & 89.10 ($\pm$ 0.7)                  & \textbf{90.56} ($\pm$ 0.4)     &0 \\
N(0,0.01)+SP+AP             & 95.60 ($\pm$ 0.0)    & 92.30 ($\pm$ 0.1)            & 92.66 ($\pm$ 0.0)                   & \textbf{92.76} ($\pm$ 0.1)     &2 \\
N(0,0.02)+No Protection     & 10.88 ($\pm$ 1.2)    & 45.36 ($\pm$ 8.8)             & 65.50 ($\pm$ 7.0)                   & \textbf{77.78} ($\pm$ 3.7)     &0 \\
N(0,0.02)+SP+AP             & 95.70 ($\pm$ 0.0)     & 91.68 ($\pm$ 0.4)            & 92.30  ($\pm$ 0.1)                   & \textbf{92.36} ($\pm$ 0.2)       &2 \\
           
\bottomrule
\\
\end{tabular}
}
\caption{Accuracy of ResNet-20 distilled from ResNet-18 on CIFAR-10 when weights are perturbed by the Gaussian noise. SP: sign protection, AP: adaptive protection (adaptive mapping+sparsity-driven adaptive redundancy). During the distillation of noisy student, a Gaussian noise with N(0,0.01) is injected onto the weights. Experiments are conducted five times.}
\label{tab:cifar_distill_Gaussian_app}
\end{table*}

\setlength{\tabcolsep}{3pt}
\begin{table*}[t]
\centering
\resizebox{0.9\textwidth}{!}{
\begin{tabular}{lccccccc}
\toprule
                            & Teacher                       & Teacher                               & Student                       & Noisy Student         & Noisy Student & Number of  & Additional  \\ 
                            & LeNet                     &  MLP                                      & MLP            & MLP                      & MLP  & PCM cells & Bits  \\
                            &                           &  (Student baseline)                    & (No Noise)            & (with N(0,0.1))              & (with N(0,0.006))          &  &  \\ \midrule
No Noise                    & 99.20                 & 97.50                                  & 97.80       & 96.30                 & \textbf{97.30}    & 16 & 0\\
PCM+SP                      & 98.94 ($\pm$ 0.0)   & 91.86 ($\pm$ 1.7)             & 87.46 ($\pm$ 3.7)      & \textbf{95.46}($\pm$ 0.1)     & 93.12 ($\pm$ 1.4)     & 1 & 1\\
PCM+AP                      & 98.60 ($\pm$ 0.2)    & 92.76 ($\pm$ 1.6)             & 95.40 ($\pm$ 1.0)     & 95.78($\pm$ 0.2)     & \textbf{96.04} ($\pm$ 0.4) &1 & 1\\
PCM+SP+AP                   & 99.20 ($\pm$ 0.0)   & 97.04 ($\pm$ 0.1)             & 97.46 ($\pm$ 0.2)      & 96.12 ($\pm$ 0.1)     & \textbf{97.58} ($\pm$ 0.1)     & 1 & 2\\
PCM+SP+AP+Sens.             & 99.20 ($\pm$ 0.0)   & 97.20 ($\pm$ 0.0)              & 97.72 ($\pm$ 0.1)      & 96.14($\pm$ 0.1)     & \textbf{97.80} ($\pm$ 0.1)      & 1 & 3 \\
PCM+AP+Sens.             & 98.88 ($\pm$ 0.1)    & 95.04 ($\pm$ 0.5)             & 97.10 ($\pm$ 0.1)      & 95.92($\pm$ 0.1)     & \textbf{97.46} ($\pm$ 0.1)      & 1 & 3 \\
       
\bottomrule
\\
\end{tabular}
}
\caption{Accuracy of MLP distilled from LeNet on MNIST when weights are perturbed by the PCM cells. SP: sign protection, AP: adaptive protection (adaptive mapping+sparsity-driven adaptive redundancy), Sens.: sensitivity-driven adaptive redundancy. Experiments are conducted five times.}
\label{tab:mnist_distill_app}
\end{table*}

\setlength{\tabcolsep}{3pt}
\begin{table*}[t]
\centering
\resizebox{0.8\textwidth}{!}{
\begin{tabular}{lccccccc}
\toprule
                            & Teacher                       & Teacher                               & Student                       & Noisy Student         & Noisy Student  & Additional  \\ 
                            & LeNet                     &  MLP                                      & MLP            & MLP                      & MLP   & Bits  \\
                            &                           &  (Student baseline)                    & (No Noise)            & (with N(0,0.1))              & (with N(0,0.006))          &  &  \\ \midrule
No Noise                    & 99.20                 & 97.50                                  & 97.80       & 96.30                 & \textbf{97.30}    &0\\
N(0,0.1)+No Protection                    & 22.60 ($\pm$ 7.7)    & 77.72 ($\pm$ 3.8)            & 58.90 ($\pm$ 3.6)      & \textbf{93.80} ($\pm$ 0.4)     & 62.20 ($\pm$ 3.5)     &0\\
N(0,0.1)+SP+AP              & 99.22 ($\pm$ 0.0)   & 95.92 ($\pm$ 0.3)             & 95.96 ($\pm$ 0.8)      & 95.82 ($\pm$ 0.2)     & \textbf{97.02} ($\pm$ 0.2)      & 2\\
N(0,0.2)+No Protection                   & 12.02 ($\pm$ 3.0)    & 33.50 ($\pm$ 3.4)             & 24.78 ($\pm$ 6.3)     & \textbf{74.82}($\pm$ 5.0)   & \textbf{25.02}($\pm$ 2.8)  & 0\\
N(0,0.2+SP+AP)             & 99.18 ($\pm$ 0.1)    & 90.36 ($\pm$ 1.0)             & 86.30 ($\pm$ 4.6)      & \textbf{95.06}($\pm$ 0.3)     & 93.06 ($\pm$ 2.1)       &2 \\
N(0,0.06)+No Protection                  & 99.24 ($\pm$ 0.0)   & 97.38 ($\pm$ 0.1)             & 97.82 ($\pm$ 0.1)        & 96.28 ($\pm$ 0.0)    & \textbf{97.86} ($\pm$ 0.0)       & 0\\
N(0,0.06)+SP+AP            & 99.20 ($\pm$ 0.0)     & 97.50 ($\pm$ 0.0)             & 97.80 ($\pm$ 0.0)        & 96.30 ($\pm$ 0.0)    & \textbf{97.90} ($\pm$ 0.0)       &2 \\
\bottomrule
\\
\end{tabular}
}
\caption{Accuracy of MLP distilled from LeNet on MNIST when weights are perturbed by the Gaussian noise. SP: sign protection, AP: adaptive protection (adaptive mapping+sparsity-driven adaptive redundancy). Experiments are conducted five times.}
\label{tab:mnist_distill_Gaussian_app}
\end{table*}

\end{document}